\documentclass[10pt,twocolumn,letterpaper]{article}

\usepackage[pagenumbers]{cvpr} 
\usepackage{booktabs}
\newcommand\mypar[1]{\par\vspace{0.5mm}\noindent\textbf{#1}\;\;}

\definecolor{cvprblue}{rgb}{0.21,0.49,0.74}
\usepackage{multicol}
\usepackage{colortbl}
\usepackage{multirow}

\usepackage{etoc}
\usepackage{minitoc}
\usepackage{bm}
\usepackage{algorithm}
\usepackage{listings}
\usepackage{amsmath,amsfonts,bm}

\usepackage{pifont}
\newcommand{\cmark}{\ding{51}}%
\newcommand{\xmark}{\ding{55}}%
\usepackage[pagebackref,breaklinks,colorlinks,allcolors=cvprblue]{hyperref}

\title{RandAR: Decoder-only Autoregressive Visual Generation in Random Orders}

\author{
Ziqi Pang$^{1*}$\quad Tianyuan Zhang$^{2*}$\quad Fujun Luan$^{3}$\quad Yunze Man$^{1}$\quad Hao Tan$^{3}$\quad Kai Zhang$^{3}$\quad \\ William T. Freeman$^{2}$\quad Yu-Xiong Wang$^{1}$ \\
{$^{1}$UIUC \quad $^{2}$MIT\quad $^3$Adobe Research}
}

\lstdefinestyle{mocov3}{
  backgroundcolor=\color{white},
  basicstyle=\fontsize{7.5pt}{7.5pt}\ttfamily\selectfont,
  columns=fullflexible,
  breaklines=true,
  captionpos=b,
  commentstyle=\fontsize{7.5pt}{7.5pt}\color[rgb]{0.25,0.5,0.5},
  keywordstyle=\fontsize{7.5pt}{7.5pt}\color[rgb]{0.85,0.18,0.50},
}

\begin{document}
\twocolumn[{%
\renewcommand\twocolumn[1][]{#1}%
\maketitle
\begin{center}
    \centering
    \captionsetup{type=figure}
    \includegraphics[width=0.99\textwidth]{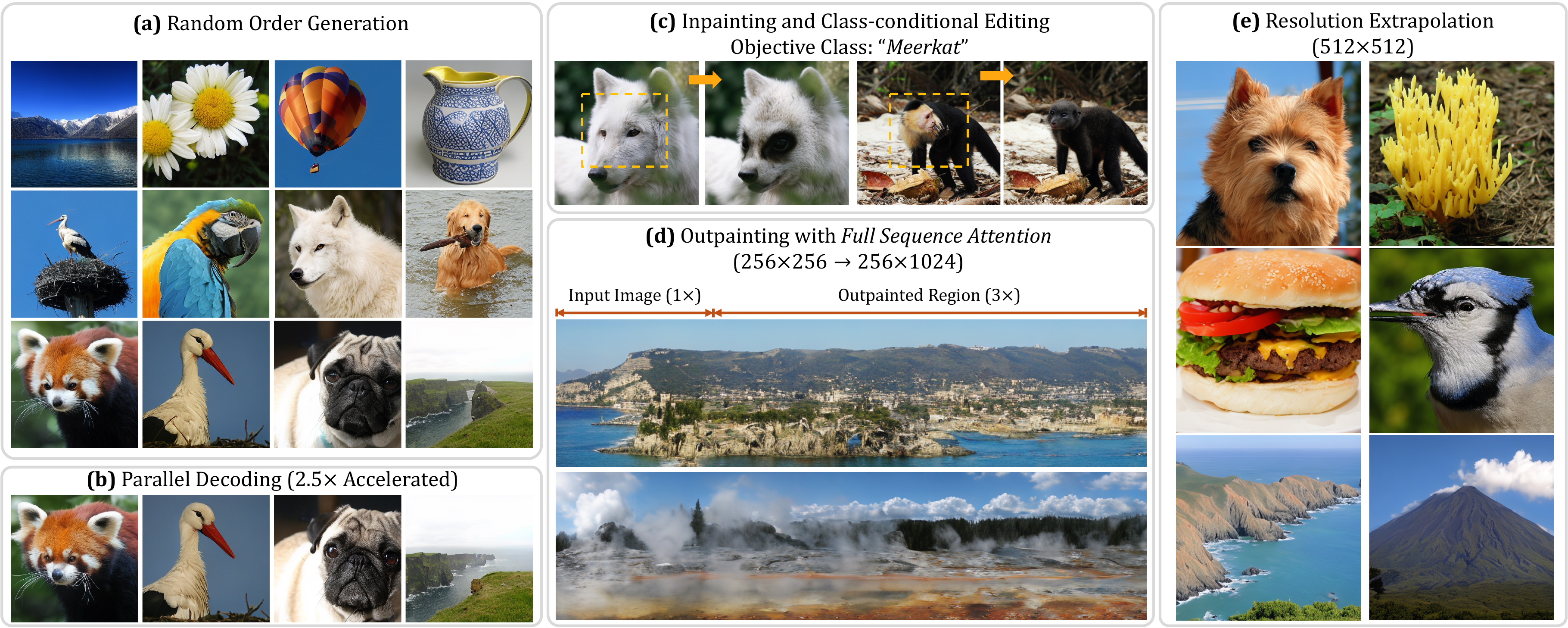}
    \vspace{-3mm}
    \captionsetup{type=figure}
    \captionof{figure}{Our \textbf{RandAR} enables GPT-style causal decoder-only transformers to generate images via \emph{random-order next-token prediction}, which entirely removes the raster-order sequencing inductive bias of previous decoder-only models. RandAR not only (a) generates images of comparable quality, but also shows multiple zero-shot capabilities, including (b) parallel decoding for acceleration, (c) inpainting, (d) outpainting, and (e) zero-shot generalization from a 256$\times$256 model to synthesize high-resolution images. Zoom in for image details. }
    \vspace{-0mm}
    \label{fig:teaser}
\end{center}
}]
\begin{abstract}
We introduce RandAR, a decoder-only visual autoregressive (AR) model capable of generating images in arbitrary token orders. Unlike previous decoder-only AR models that rely on a predefined generation order, RandAR removes this inductive bias, unlocking new capabilities in decoder-only generation. Our essential design enables random order by inserting a ``position instruction token'' before each image token to be predicted, representing the spatial location of the next image token. Trained on randomly permuted token sequences -- a more challenging task than fixed-order generation, RandAR achieves comparable performance to its conventional raster-order counterpart. More importantly, decoder-only transformers trained from random orders acquire new capabilities. For the efficiency bottleneck of AR models, RandAR adopts parallel decoding with KV-Cache at inference time, enjoying 2.5$\times$ acceleration without sacrificing generation quality.
Additionally, RandAR supports inpainting, outpainting and resolution extrapolation in a zero-shot manner.
We hope RandAR inspires new directions for decoder-only visual generation models and broadens their applications across diverse scenarios. Our project page is at \href{https://rand-ar.github.io/}{https://rand-ar.github.io/}.

\end{abstract}    
\begin{NoHyper}
\def\thefootnote{*}\footnotetext{Equal Contribution.}
\end{NoHyper}
\def\thefootnote{\arabic{footnote}}
\section{Introduction}
\label{sec:intro}

Inspired by the success of ``next-token prediction'' in language modeling, computer vision researchers have explored using GPT-style \emph{uni-directional decoder-only} transformers for image generation. The typical approach tokenizes an image into discrete 2D tokens, arranges them into 1D sequences in a row-major (raster) order from top-left to bottom-right, and applies a decoder-only transformer for sequential next visual token prediction. This design has shown promising results in uni-modal and multi-modal image generation~\cite{yu2024randomized, sun2024autoregressive, wang2024emu3, team2024chameleon, zhou2024transfusion}. However, enforcing a uni-directional raster order limits the decoder-only transformers from modeling the \emph{bi-directional} context in 2D images -- a constraint that their encoder-decoder counterparts, \emph{e.g.}, MaskGIT~\cite{chang2022maskgit} and MAR~\cite{li2024autoregressive}, do not face. Fundamental questions thus remain: Is pre-defined raster-order sequencing truly a necessary and useful inductive bias for decoder-only image generators? If not, how can we equip these models with bi-directional modeling capabilities?

To this end, we propose \emph{random-order next-token prediction}, termed ``\textbf{RandAR}'', which enables \emph{fully randomized generation order} during \emph{both training and inference} time. This approach brings the encoder-decoder models' advantage of bi-directional context modeling to decoder-only models, while preserving the plain transformer architecture, simple next-token prediction mechanism, and KV-Cache acceleration.

Concretely, RandAR arranges 2D image tokens in a random-order 1D sequence, with specially designed \emph{positional instruction tokens} inserted before each image token to indicate their spatial locations. Then, we apply a standard uni-directional decoder-only transformer for next-token prediction. Such random ordering encourages the model to learn non-local correlations. Although this setup is more challenging -- introducing 256$!$ possible sequences permutations for 256-token 256$\times$256 images -- we demonstrate that RandAR achieves comparable generation quality to its raster-order counterpart on the ImageNet benchmark~\cite{deng2009imagenet} (examples in Fig.~\ref{fig:teaser}\textcolor{cvprblue}{(a)}).

More importantly, introducing random order to uni-directional decoder-only models unleashes their critical new \emph{zero-shot} capabilities. As a direct advantage, RandAR inherently supports \emph{parallel decoding} with no post-training required, accelerating sampling speed by 2.5$\times$ without compromising quality  (Fig.~\ref{fig:teaser}\textcolor{cvprblue}{(b)}). Furthermore, random-order prediction provides the decoder-only model with a level of flexibility exceeding that of raster-order models, thereby unlocking new applications. Beyond inpainting~\cite{chang2022maskgit} (Fig.~\ref{fig:teaser}\textcolor{cvprblue}{(c)}), RandAR can conduct zero-shot outpainting with a single round of \emph{full-sequence attention} on an extrapolated number of tokens, leading to highly consistent details and patterns (Fig.~\ref{fig:teaser}\textcolor{cvprblue}{(d)}). \emph{Surprisingly}, we demonstrate that RandAR, trained on 256$\times$256 resolution, can synthesize 512$\times$512 images by leveraging specially designed generation orders with full-sequence attention (Fig.~\ref{fig:teaser}\textcolor{cvprblue}{(e)}). Unlike conventional sliding-window outpainting, our high-resolution images exhibit unified objects with richer details. Finally, we show that RandAR's causal transformer can directly extract bi-directional features by processing image tokens twice, while raster-order models struggle with such generalization.

To summarize, our contributions are:
\begin{enumerate}
    \item We introduce \textbf{RandAR}, a framework enabling causal decoder-only models to conduct random-order next-token prediction.
    \item We validate our design on the ImageNet benchmark, demonstrating comparable generation quality to raster-order counterparts while reducing the inference latency by 2.5 times through parallel decoding.
    \item RandAR unlocks a wide range of zero-shot capabilities: inpainting, bi-directional feature extraction, and full attention on extrapolated sequence lengths for outpainting and resolution extrapolation.
\end{enumerate}
We hope RandAR removes a significant barrier to modeling 2D images with uni-directional decoder-only transformers and inspires further exploration of its broader capabilities.

\section{Related Work}
\label{sec:related}
\begin{figure*}
    \centering
    \includegraphics[width=0.9\linewidth]{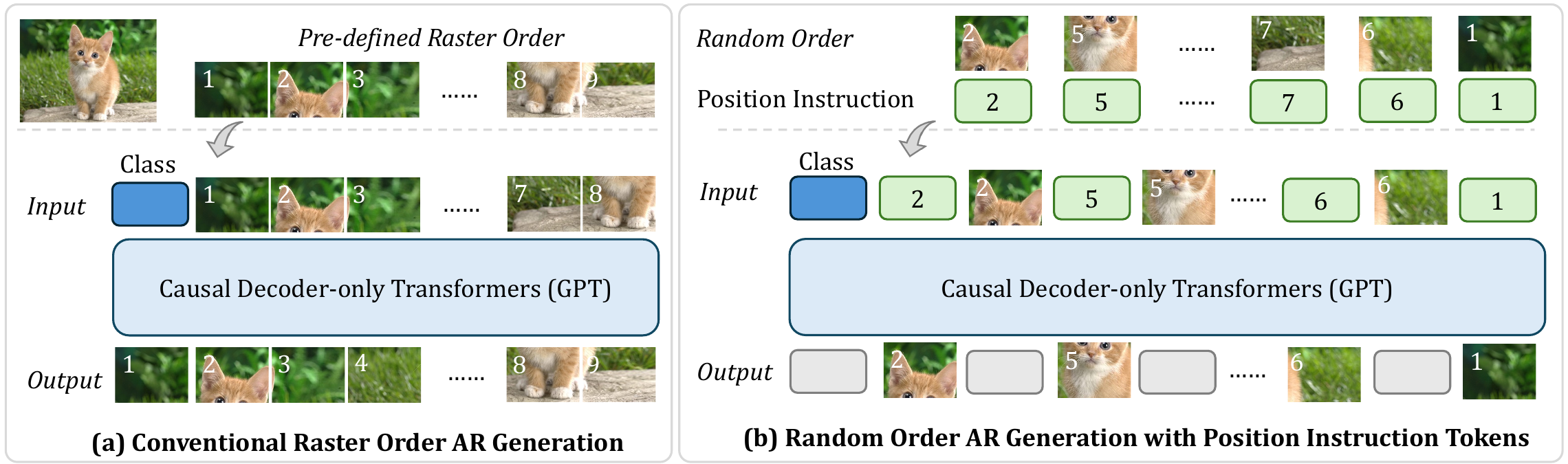}
    \vspace{-3mm}
    \caption{Overview of our \textbf{RandAR}. \textbf{(a)} Conventional autoregressive generation typically enforces a fixed order, \emph{e.g.,} raster order, allowing the model to memorize token orders. 
    \textbf{(b)} Our RandAR enables random order generation by inserting a \emph{position instruction token} before each image token to be predicted. This design seamlessly integrates with the next-token prediction framework using decoder-only transformer.}\vspace{-3mm}
    \label{fig:method} 
\end{figure*}

\mypar{AR Language Generation.} Current large language models (LLMs) generate text as 1D sequence autoregressively. Since the initial efforts of scaling up, two distinct architectures have emerged: bi-directional BERT-like models~\cite{kenton2019bert, yang2019xlnet, liu2019roberta, raffel2020exploring} and unidirectional GPT-like models~\cite{touvron2023llama, brown2020language, radford2018improving, touvron2023llama2, groeneveld2024olmo}. The BERT-like architecture follows an encoder-only or encoder-decoder design and usually uses mask tokens as placeholders for language generation. In comparison, the GPT-like architectures are plain decoder-only transformers in causal attention, which learn to conduct ``next-token prediction''. Decoder-only architectures have recently become the dominant choice for language generation due to simplicity, scalability, and zero-shot generalization across various tasks. Inspired by the versatility of GPT models, we aim to build on the current decoder-only \emph{image} model, reducing its inductive bias and expanding its zero-shot capabilities by adopting random 2D generation orders.

\mypar{Decoder-only AR Image Generation.} Models~\cite{wang2024emu3, zhou2024transfusion, xie2024show, team2024chameleon, ramesh2021zero, yu2024randomized} represented by VQGAN~\cite{esser2021taming}, RQTran~\cite{lee2022autoregressive}, and LLaMAGen~\cite{sun2024autoregressive} directly transfer the GPT-style decoder-only language models for visual generation. These models turn 2D images into 1D sequences following a pre-defined factorization, typically raster order or coarse-to-fine resolutions modeled by bi-directional attention~\cite{tian2024visual}. Instead, our RandAR provides a simple strategy empowering decoder-only transformers for arbitrary generation orders, which greatly extends their capabilities.

\mypar{Masked AR Image Generation.} Masked AR methods~\cite{li2023mage, li2024autoregressive, chang2022maskgit, gao2023masked, luo2024open, yu2023magvit, chang2023muse, yu2024image, weber2024maskbit, fan2024fluid, liu2024customize, liu2024solving} employ bi-directional attention commonly implemented with an encoder-decoder design, learning to decode place-holding mask tokens in random orders. While these architectures lack KV-Cache support and direct compatibility with large language models (LLMs), \emph{e.g.}, MaskGIT~\cite{chang2022maskgit} and MAR~\cite{li2024autoregressive}, they offer greater flexibility and versatility than raster-order decoder-only models, such as parallel decoding and image inpainting. Therefore, the major objective of our paper is to introduce such random order and bi-directional ability into decoder-only models via our RandAR, bridging the conceptual gap between unidirectional decoder-only image generation and masked image generation.

\section{Method}
\label{sec:method}
\subsection{Preliminaries}
\label{sec:prelim}

\noindent\textbf{Decoder-only Autoregressive Models}
generate sequences by predicting each token sequentially, using only past information.
Formally, given a 1D sequence of $N$ variables, denoted as $\mathbf{x}=[x_1, x_2, ..., x_{N}]$,  an autoregressive model is trained to model the probability distribution of each variable $x_n$ based on its precedents $[x_1, .., x_{n-1}]$:
\begin{equation}
\label{eqn:autoregressive}
    p_{\theta}(\mathbf{x})=\prod_{n=1}^{N}p_{\theta}(x_n|x_1, ..., x_{n-1}),
\end{equation}
where $p_{\theta}(\mathbf{x})$ may be implemented using a multinomial distribution for discrete tokens or a diffusion model for continuous tokens \cite{li2024autoregressive}. Currently, one of the most scalable implementations of autoregressive models employs a stack of unidirectional transformer layers with causal attention, i.e., decoder-only transformer~\cite{brown2020language}. 

To apply this unidirectional approach to image generation, 2D images must be converted to 1D sequences. Existing works~\cite{van2016conditional, esser2021taming, sun2024autoregressive, li2024autoregressive} enforce a predefined generation order, such as the raster-line order. This design introduces an inductive bias, focusing the network on predicting adjacent patches, only using context from one direction. Models trained in this way are limited to fixed generation orders and lack flexibility for tasks such as inpainting and outpainting.

In contrast, RandAR removes this inductive bias entirely by generating image token sequences in arbitrary orders. Building upon prior decoder-only visual autoregressive models~\cite{sun2024autoregressive}, we introduce minimal modifications (only one additional trainable parameter) to support random-order next-token prediction. Furthermore, we show that our model achieves generation quality comparable to raster-order models under fair comparison, despite the increased complexity of learning across  $(N!)$ possible orders. 

\subsection{RandAR Framework}
\label{sec:framework}

Our goal is to introduce minimal modifications to the original GPT-style visual autoregressive framework~\citep{sun2024autoregressive} to enable random order generation. The key insight is that the model needs to be informed about the position of each next token. Our solution is straightforward: we insert a special token, called  \emph{position instruction token}, before each image token to be predicted, to represent its position. Specifically, we arrange image tokens in raster order, then randomly shuffle the sequence and drop the last: 
\begin{equation}
    [x_{1}^{\pi(1)}, x_{2}^{\pi(2)}, \ldots, x_{N-1}^{\pi(N-1)}],
    \label{eq:random_order}
\end{equation}
where $x_{i}^{\pi(i)}$ is the $i$-th token in this randomly shuffled sequence of length $N$, and ${\pi(i)}$ denotes its original position in raster order. We then insert a positional instruction token $P_{i}^{\pi(i)}$ before each image token $x_{i}^{\pi(i)}$, as Fig.~\ref{fig:method}:
\begin{equation}
    [P_{1}^{\pi(1)}, x_{1}^{\pi(1)}, P_{2}^{\pi(2)}, x_{2}^{\pi(2)}, \ldots, x_{N-1}^{\pi(N-1)}, P_{N}^{\pi(N)}].
    \label{eq:random_order_position}
\end{equation}

RandAR then applies a standard decoder-only transformer with causal attention to this sequence and supervises the prediction of each position instruction token with its subsequent image token. This random-order autoregressive modeling can be formalized as:
\begin{equation}
\label{eqn:position_instruction}
p_{\theta}(\mathbf{x}|\mathbf{P}) = \prod_{n=1}^{N} p_{\theta}(x_{n}^{\pi(n)}|P_{1}^{\pi(1)}, x_{1}^{\pi(1)}, \ldots, x_{n-1}^{\pi(n-1)}, P_{n}^{\pi(n)}).
\end{equation}
For simplicity, we omit the subscript ${\pi(i)}$ in later equations. 
Adding position instruction tokens before each image token resembles the concept of target-aware representation, as discussed in XLNet~\cite{yang2019xlnet}.

\mypar{Position Instruction Tokens.}  We insert \emph{position instruction tokens}, representing the spatial location of each next token, to enable random-order generation, shown in Fig.~\ref{fig:method}\textcolor{cvprblue}{(b)}. 
For each position, we use a shared learnable embedding \( \boldsymbol{e} \) “rotated” with the 2D coordinates of the next image token to be predicted, following 2D-RoPE~\cite{su2024roformer}. The position instruction token for an image token at position  $(h_i, w_i)$ is:
\begin{equation}
    P_i = \mathrm{RoPE}(\boldsymbol{e}, h_i, w_i).
\end{equation}
The ordinary RoPE~\cite{su2024roformer} is a relative positional embedding only effective inside the attention operator.
We empirically find that it also works well as a global position embedding in our position instruction design.
Alternative designs, such as dense learnable positional embeddings or merging position instruction tokens with image tokens, are examined in the ablation study in Sec.~\ref{sec:main_results}.

\mypar{Architecture.} RandAR follows the architecture of LLaMAGen~\cite{sun2024autoregressive}, using a stack of decoder-only transformers with 2D RoPE~\cite{su2024roformer} as relative positional encoding within the attention module. For class-to-image generation on ImageNet~\cite{deng2009imagenet}, class IDs are embedded as learnable embeddings. Only one trainable embedding for position instruction tokens is added beyond LLaMAGen~\cite{sun2024autoregressive} to support random order next token prediction.

\mypar{Training.} We train RandAR with random sequence orders sampled from all $(N!)$ possible permutations. Using the tokenizer from LLaMAGen~\cite{sun2024autoregressive}, which tokenizes a 256$\times$256 image to  $N$=16$\times$16 2D discretized tokens, this leads to approximately $256!=8\times10^{506}$ possible orders. Although training on ImageNet~\cite{deng2009imagenet} for 300 epochs only covers a small number of $3\times 10^8$ orders at most, RandAR learns the ability to generate images in random orders.

\mypar{Inference.} Given an arbitrary order for inference, we first compute the corresponding position instruction tokens, then iteratively sample the predicted image tokens with standard next token prediction. We discover that RandAR trained with random orders generates better images with random sequence orders at the inference time than raster orders. More analysis on inference orders is in Table~\ref{table:supp_generation_orders}.

\subsection{RandAR Enables Parallel Decoding}
\label{sec:parallel_decoding}

Decoder-only AR image models, by default, generate one token at a time during inference.
However, this sequential decoding is bottlenecked by hardware's memory bandwidth~\cite{shazeer2019fast, kimsqueezellm, cai2024medusa} (also well-known as ``memory wall''), as each new token generation step requires a forward pass through the model, and the model needs to load all parameters into GPU registers, which is a process considerably slower than computation. Therefore, the number of steps is a crucial factor for the latency of AR models.

Fortunately, RandAR can predict tokens at any location based on previously generated tokens. This enables parallel decoding, where RandAR simultaneously predicts tokens at multiple locations in one iteration. By reducing the number of forward steps, parallel decoding significantly decreases generation latency.

We illustrate two-token parallel decoding as an example.
Suppose the generated token and position instruction token sequence up to now is
$\mathbf{x}_{1:n-1} = [P_1, x_1, ..., P_{n-1}, x_{n-1}]$,
at the new iteration, we append two position instruction tokens $[P_n, P_{n+1}]$ at the end of the sequence:
\begin{equation}
    [P_1, x_1, ..., P_{n-1}, x_{n-1}, P_n, P_{n+1}],
\end{equation}
and pass it through the network. 
RandAR then predicts and sample the next two tokens $[x_n, x_{n+1}]$. After sampling, we rearrange the newly added sequence to the training-time interleaved format as follows: 
\begin{equation}
    [P_1, x_1, ..., P_{n-1}, x_{n-1}, P_n, x_n, P_{n+1}, x_{n+1}].
\end{equation}
In subsequent iterations, we append new position instruction tokens at the sequence’s end and repeat this process. Such rearrangement ensures the sequence maintains the same interleaved format used during training, with each image token preceded by a position instruction token.

Our parallel decoding requires no training modification or fine-tuning. It preserves causal masking and remains compatible with the KV cache. Such parallel decoding is already explored in masked AR methods like MaskGIT~\cite{chang2022maskgit, li2024autoregressive}, but lacks the support of KV-cache acceleration. We show our effective acceleration ratio with parallel decoding in Table~\ref{tab:latency}.

\begin{figure*}
    \centering
    \includegraphics[width=0.95\linewidth]{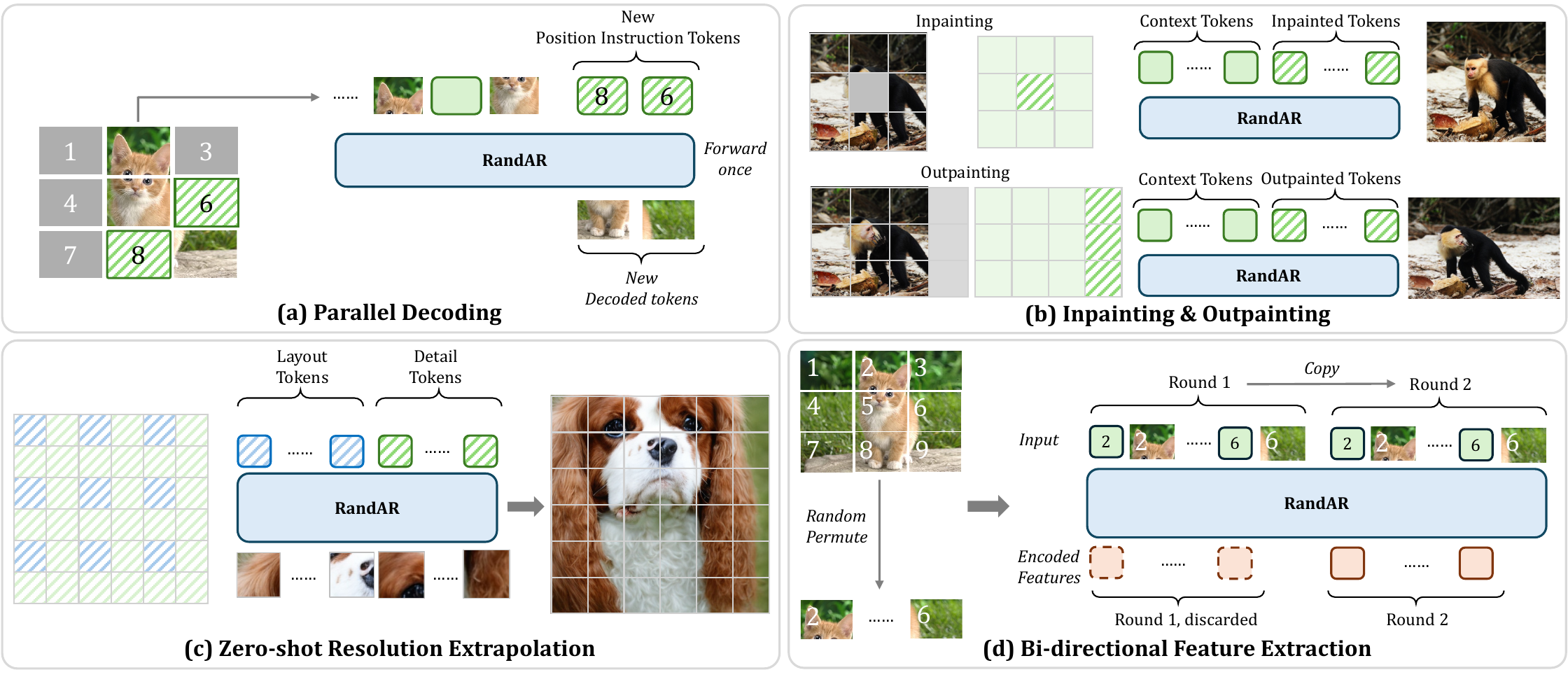}
    \vspace{-3mm}
    \caption{Zero-shot capabilities of RandAR. \textbf{(a)} RandAR directly enables parallel decoding to accelerate AR generation (Sec.~\ref{sec:parallel_decoding}). \textbf{(b)} Without the order constraint, our RandAR can support inpainting (Sec.~\ref{sec:inpainting}) and outpainting (Sec.~\ref{sec:outpainting}). \textbf{(c)} RandAR trained on 256$\times$256 can also zero-shot generalize to synthesize 512$\times$512 higher-resolution images with customized order (Sec.~\ref{sec:resolution}). \textbf{(d)} The random order training enables to extract features with bi-directional contexts, \emph{i.e.}, the output from the 2nd pass of tokens (Sec.~\ref{sec:bidirectional_encoding}). }\vspace{-2mm}
    \label{fig:applications}
\end{figure*}

\subsection{Zero-shot Applications for RandAR}
\label{sec:case_studies}

\subsubsection{Inpainting and Class-conditional Image Editing}
\label{sec:inpainting}

A predefined generation order limits AR image generators in image manipulation tasks, as they cannot aggregate contextual information from different parts of the image. Consequently, decoder-only AR models, especially raster-order ones, lack zero-shot capability for tasks like inpainting and class-conditioned image editing, which are achievable with masked image modeling methods such as MaskGIT~\cite{chang2022maskgit}.  

RandAR overcomes this limitation by enabling unidirectional transformers to incorporate contextual information from any part and any direction of the image. For inpainting and class-conditional image editing, we simply position visible image tokens and their corresponding position instruction tokens before the instruction tokens for the areas to be edited. RandAR then completes the remaining tokens autoregressively,  as in Fig.~\ref{fig:applications}\textcolor{cvprblue}{(b)}. The capability to use spatially randomized context and support arbitrary sampling orders is necessary for these tasks and is also central to RandAR’s functionality. Results are in Sec.~\ref{sec:exp_inpainting}.

\subsubsection{Outpainting}
\label{sec:outpainting}
Outpainting requires extending the content of an existing image beyond its boundary in a visually coherent and contextually relevant way. Raster-order models can only extract contextual information from top-left image patches to predict the next token; thus, they have to employ strategies like sliding window as in VQGAN~\cite{esser2021taming} and can only take partial contexts into account. In contrast, our RandAR can directly process all the image tokens from the conditional image, as in Fig.~\ref{fig:inpainting}, where we outpaint the original 256$\times$256 image to 256$\times$1024 by extending it threefold. When outpainting beyond the training context length, we extrapolate RoPE to the target length, then use full sequence attention to model all the tokens jointly. For comparison, we also display results with sliding window attention using RandAR in the bottom row of Fig.~\ref{fig:inpainting} (Sec.~\ref{sec:exp_outpainting}).

\subsubsection{Resolution Extrapolation}
\label{sec:resolution}
Unlike outpainting, which extends an image’s boundaries, resolution extrapolation requires generating finer details within the existing image boundaries — essentially \emph{outpainting in the frequency domain}.  We explore this task by generating 512$\times$512 images with RandAR trained solely on 256$\times$256 ImageNet images, without additional fine-tuning. Our resolution extrapolation involves two steps as illustrated in Fig.~\ref{fig:applications}\textcolor{cvprblue}{(c)}: (1) Generating tokens at even coordinates to establish the overall layout, using interpolated RoPE positional embeddings; (2) Generating the tokens at the remaining coordinates. In this step, we use top-k high-frequency components in the extrapolated RoPE to replace the interpolated RoPE, which better captures the finer details, motivated by NTK-RoPE~\cite{fu2024data} for extending context lengths in language models. 

This hierarchical decoding schedule relies on RandAR’s ability to process tokens in random orders. For comparison, we also show generated results without this decoding approach, which exhibit visual inconsistencies. Inspired by classifier-free guidance~\cite{ho2022classifier}, we introduce a new type of guidance: \emph{spatial contextual guidance} (SCG) to enhance high-frequency details in resolution extrapolation. At inference, we maintain a secondary sequence where each newly sampled token is randomly dropped with a $25\%$ probability. We then combine predicted logits from both the original and token-dropped sequences to sharpen the final output. Additional details and results are provided in Sec.~\ref{sec:scg}.

\begin{table*}[h]
\centering
\caption{Model comparisons on class-conditional ImageNet 256×256 benchmark. Metrics are Fréchet inception distance (FID), inception score (IS), precision and recall. ``↓'' or ``↑'' indicate lower or higher values are better. ``-re'' means using rejection sampling. $^*$ represents training at 384x384 resolution, and resized to 256x256 for evaluation.  The raster-order counterpart is trained using the same architecture and setup as our RandAR for a fair comparison. RandAR is the only decoder-only method capable of generating images in random token orders, and it achieves comparable performance to the raster-order counterpart despite learning a more challenging task of 256! orders. }\vspace{-2mm}
\label{table:main_results}
\resizebox{0.77\linewidth}{!}{
\begin{tabular}{c|l|c|c|c|c|c|c}
\toprule
\textbf{Type} & \textbf{Model} & \textbf{\#Para.} & \textbf{FID↓} & \textbf{IS↑} & \textbf{Precision↑} & \textbf{Recall↑}  & \textbf{Steps}\\
\midrule
GAN & BigGAN \cite{brock2018large} & 112M & 6.95 & 224.5 & 0.89 & 0.38 & 1\\
& GigaGAN \cite{kang2023scaling} & 569M & 3.45 & 225.5 & 0.84 & 0.61 & 1 \\
& StyleGan-XL \cite{sauer2022stylegan} & 166M & 2.30 & 265.1 & 0.78 & 0.53 & 1 \\
\midrule
Diffusion & ADM \cite{dhariwal2021diffusion} & 554M & 4.59 & 186.70 & 0.82 & 0523 & 250 \\
& LDM-4 \cite{rombach2022high} & 400M & 3.60 & 247.7 & -- & -- & 250\\
& DiT-XL \cite{peebles2023scalable} & 675M & 2.27 & 278.2 & 0.83 & 0.57 & 250\\
& SiT-XL \cite{ma2024sit} & 675M & 2.06 & 270.3 & 0.82 & 0.59 & 250 \\
\midrule
Bi-directional AR & MaskGIT-re \cite{chang2022maskgit} & 227M & 4.02 & 355.6 & -- & -- & 8 \\
& MAGVIT-v2 \cite{yu2023language} & 307M & 1.78 & 319.4 & -- & -- & 64 \\
& MAR-L\cite{li2024autoregressive} & 479M & 1.98 & 290.3 & -- & -- & 64 \\
& MAR-H\cite{li2024autoregressive} & 943M & 1.55 & 303.7 & 0.81 & 0.62 & 256 \\
& TiTok-S-128~\cite{yu2024image} & 287M & 1.97 & 281.8 & -- & -- & 64 \\
\midrule
Casual AR & VQGAN-re \cite{esser2021taming} & 1.4B & 5.20 & 280.3 & -- & -- & 256  \\
& RQTran.-re \cite{lee2022autoregressive} & 3.8B & 3.80 & 323.7 & -- & -- & 64 \\
& VAR \cite{tian2024visual} & 600M & 2.57 & 302.6 & 0.83 & 0.56 & 10 \\
& VAR \cite{tian2024visual} & 2.0B & 1.92 & 350.2 & 0.82 & 0.59 & 10 \\
& SAR-XL \cite{liu2024customize} & 893M & 2.76 & 273.8 & 0.84 & 0.55 & 256 \\
& RAR-B \cite{yu2024randomized} & 261M & 1.95 & 290.5 & 0.82 & 0.58 & 256 \\
& RAR-L \cite{yu2024randomized} & 461M & 1.70 & 299.5 & 0.81 & 0.60 & 256 \\
& RAR-XXL \cite{yu2024randomized} & 955M & 1.50 & 306.9 & 0.80 & 0.62 & 256 \\
& RAR-XL \cite{yu2024randomized} & 1.5B & 1.48 & 326.0 & 0.80 & 0.63 & 256 \\
& Open-MAGVIT2-XL \cite{luo2024open} & 1.5B & 2.33 & 271.8 & 0.84 & 0.54 & 256\\
& LlamaGen-L\cite{sun2024autoregressive}  & 343M & 3.07 & 256.06 & 0.83 & 0.52 & 256 \\
& LlamaGen-XL$^*$\cite{sun2024autoregressive} & 775M & 2.62 & 244.08 & 0.80 & 0.57 & 576 \\
& LlamaGen-XXL$^*$\cite{sun2024autoregressive}  & 1.4B & 2.34 & 253.90 & 0.80 & 0.59 & 576 \\
& LlamaGen-3B$^*$\cite{sun2024autoregressive}  & 3.1B & 2.18 & 263.33 & 0.81 & 0.58 & 576 \\
\midrule
Casual AR & Raster-order Counterpart & 343M & 2.20 & 274.26  & 0.80 & 0.59 & 256 \\
& Raster-order Counterpart & 775M & 2.16 &  282.71 & 0.80  & 0.61 & 256 \\
\midrule
Casual AR & RandAR-L & 343M & 2.55 & 288.82
 & 0.81 & 0.58 & 88\\
& RandAR-XL & 775M & 2.25 & 317.77 & 0.80 & 0.60 & 88 \\ 
& RandAR-XL & 775M & 2.22 & 314.21 & 0.80 & 0.60 & 256 \\ 
& RandAR-XXL & 1.4B & 2.15 & 321.97 & 0.79 & 0.62 & 88 \\
\bottomrule
\end{tabular}
}
\end{table*}

\subsubsection{Bi-directional Encoding}
\label{sec:bidirectional_encoding}

We find that RandAR can effectively extract features using \emph{bi-directional context} without additional training, outperforming the representations encoded by raster-order models.
Formally, an image is first tokenized and arranged into a sequence of image tokens $\mathbf{x} = [x_1, x_2, ..., x_N]$, following raster order for raster-order models and random order for random-order models.  Position instruction tokens are inserted before each image token, creating an interleaved sequence, which is then passed to the model $p_{\theta}$ as: $p_{\theta}(P_1, x_1, P_2, x_2, ..., P_N, x_N)$. The output hidden embeddings corresponding to the position instruction tokens are treated as the extracted features for the image. In a unidirectional model, earlier tokens are restricted from receiving information from later parts of the image. To enable bi-directional information aggregation with a causal transformer, we pass the sequence through the model twice:
\begin{equation}
\label{eqn:bidirectional_representation}
p_{\theta}(\underbrace{P_1, x_1, ..., P_N, x_N}_{\text{Round 1}}, \underbrace{P_1, x_1, ..., P_N, x_N}_{\text{Round 2}}),
\end{equation}
and use the features from the second round as shown in Fig.~\ref{fig:applications}\textcolor{cvprblue}{(d)}. Notably, \emph{only RandAR trained with random orders benefits from this second pass}, whereas raster-order models do not exhibit similar capabilities, as shown in Sec.~\ref{sec:exp_bidirectional_encoding}. Moreover, RandAR leverages bi-directional context in a zero-shot manner, even though the sequence lengths and formats in Eqn.~\ref{eqn:bidirectional_representation} differ significantly from its original training setups. For experimental results on ImageNet linear probing and zero-shot semantic image correspondence, please see Sec.~\ref{sec:exp_bidirectional_encoding} and Table~\ref{tab:feature_encoding}.
\section{Experiments}

\subsection{Implementation Details}

We evaluate RandAR on class-conditional image generation using the ImageNet benchmark. RandAR is implemented based on LLaMAGen~\cite{sun2024autoregressive} for standardized comparison, with only one additional trainable embedding added for position instruction tokens.

\mypar{Image Tokenizer.} We use the VQGAN~\cite{esser2021taming} tokenizer trained by LLaMAGen on ImageNet~\cite{deng2009imagenet}, which  downsamples images by 16$\times$ and has vocabulary size of 16,384.

\mypar{Decoder-only Transformer.} RandAR follows the standard design in LLaMA~\cite{touvron2023llama} with RMSNorm~\cite{zhang2019root} for normalization, SwiGLU~\cite{shazeer2020glu} for activation functions, and 2D RoPE~\cite{su2024roformer} for relative positional embeddings. The model architectures are shown in Table~\ref{tab:model_scale}.

\mypar{Training.} 
\label{sec:training_details}
The model is trained with a batch size of 1024 for 300 epochs (360K iterations) without exponential moving average. We use AdamW~\cite{loshchilov2018decoupled, kingma2014adam} optimizer with $(\beta_1, \beta_2)$ as $(0.9, 0.95)$ and the weight decay as $0.05$.   The learning rate remains constant as $4\times10^{-4}$ for the first 250 epochs without warmup, then linearly decays to  $1 \times 10^{-5}$ over the last 50 epochs. A standard token dropout of 0.1 is applied.  To support Classifier-free guidance~\cite{ho2022classifier}, class embedding is randomly dropped with a $10\%$ probability in training. 

\mypar{Inference.} The RandAR trained with random orders generates images following fully randomized orders. By default, we use 88 steps to generate 256 tokens for a 256x256 resolution image. Following MaskGIT~\cite{chang2022maskgit} and MAR~\cite{li2024autoregressive}, we apply a cosine schedule for the parallel decoding step size and a linear schedule for classifier-free guidance~\cite{ho2022classifier}.

\begin{table}
    \centering
\caption{The design choices for \emph{position instruction tokens}. Our default design uses a single shared learnable embedding with 2D RoPE~\cite{su2024roformer} for all image locations. We compare this with: (1) \emph{Dense} learnable embeddings (256 unique embeddings for 16$\times$16 tokens in 256$\times$256 resolution images); (2) \emph{Merge} position instruction tokens to its precedent image tokens by adding them together, reducing sequence length by half. However, this ``\emph{Merge}'' design reduces performance significantly when parallel decoding is applied, so we report its results without parallel decoding.}\label{tab:ablation_pe}
\vspace{-1mm}
\resizebox{0.82\linewidth}{!}{
    \begin{tabular}{l|c c c }
    \toprule
       & FID↓ & Inception Score ↑ & \#Steps \\
         \midrule
    RandAR & 2.82   & 293.6 & 88  \\
    \midrule
    Dense &  3.07  & 290.6 & 88  \\
    Dense \& Merge & 3.37   & 307.6  & 256  \\
    \bottomrule
    \end{tabular}
    }
\end{table}

\subsection{Main Results}
\label{sec:main_results}
We use Fréchet Inception Distance (FID)~\cite{heusel2017gans} as our primary metric, sampling 50,000 images with a fixed random seed and evaluating FID using code from ADM~\cite{dhariwal2021diffusion}. Following LLamaGen~\cite{sun2024autoregressive}, we also report Inception Score (IS)~\cite{salimans2016improved}, Precision and Recall~\cite{kynkaanniemi2019improved}. 

For a fair comparison, we create a raster-order counterpart using the identical setup. For all the experiments, we sweep the optimal weight for classifier-free guidance. We compare our results with current state-of-the-art methods and the raster-order counterparts in Table~\ref{table:main_results}. Results show that the XL-sized random order model reaches comparable performance with the raster counterpart,  despite the increased difficulty of random-order generation.

We also plot FID over training iterations for different model sizes in Fig.~\ref{fig:scaling_pd}\textcolor{cvprblue}{(a)}, showing consistent improvements with larger models and longer training.

\mypar{Ablation Study: Design Choices of Position Instruction Tokens.}
To enable random-order generation, we insert a \emph{position instruction token} before each image token to be predicted. Our default design uses a single learnable embedding with 2D RoPE~\cite{su2024roformer} to represent all image locations, adding only one additional learnable embedding to the existing decoder-only visual AR model. We explore two additional design choices. \emph{Dense}: A unique learnable embedding is used for each spatial location, resulting in 256 position instruction tokens for a 256$\times$256 resolution. \emph{Merge}: The position instruction token is added with the image token before it, \emph{i.e.}, each image token directly incorporates the position of the next token to be predicted. 

We train an XL-sized model with 775M parameters for each design choice over 100k iterations, following the setup in Sec.~\ref{sec:training_details}. We report FID-50K and Inception Score for each in Table~\ref{tab:ablation_pe}. Our default design shows the best performance.  Notably, the \emph{Merge} design shows degraded performance when using parallel decoding directly.

\begin{table}[t]
\centering
\caption{Ablation studies on \emph{generation orders} for RandAR. We experiment with the default fully-randomized orders, the partially randomized orders guided by priors, and the fixed orders explored in VQGAN~\cite{esser2021taming}. We discover that the default random order performs the best, and the orders of ``hierarchical random'' and ``subsample'' also outperform other orders. These indicate that RandAR benefits from the overall image contexts provided by more divergent token locations at initial steps.}\vspace{-1mm}
\label{table:supp_generation_orders}
\resizebox{0.99\linewidth}{!}{
\begin{tabular}{l|c|c|c|c|c}
\toprule
\textbf{Order} & \textbf{FID↓} & \textbf{IS↑} & \textbf{Precision↑} & \textbf{Recall↑}  & \textbf{Steps}\\
\midrule
Random & 2.25 & 317.8 & 0.80 & 0.60 & 88 \\ 
\midrule
Hierarchical Random & 2.36 & 310.2 & 0.80 & 0.60 & 88 \\
Center-first Random & 2.97 & 262.8 & 0.76 & 0.63 & 88 \\
Border-first Random & 2.56 & 300.5 & 0.78 & 0.61 & 88 \\
\midrule
Raster & 4.82 & 299.2 & 0.71 & 0.60 & 256 \\ 
Spiral-in & 3.36 & 280.1 & 0.72 & 0.64 & 256 \\
Spiral-out & 3.79 & 239.5 & 0.73 & 0.65 & 256 \\
Z-curve & 4.00 & 317.3 & 0.73 & 0.60 & 256 \\
Subsample & 2.40 & 298.8 & 0.79 & 0.61 & 256 \\
Alternate & 4.29 & 307.2 & 0.72 & 0.58 & 256 \\
\bottomrule
\end{tabular}
}
\end{table}

\mypar{Ablation Study: Inference-time Orders.} RandAR can utilize arbitrary generation orders by training on random generation orders. In addition to the row-major raster order, we investigate the generation orders from VQGAN~\cite{esser2021taming}: spiral-in, spiral-out, z-curve, subsample, and alternate. Because of the random-order generation ability of RandAR, we further propose some prior knowledge for using \emph{partially} random orders: (1) Hierarchical, which is used for our resolution extrapolation and first generates the tokens at even coordinates for a global layout; (2) Center-first, which first generates the tokens at the center 1/2$\times$1/2 of the image; (3) Border-first, which first generates the background before the center tokens. The visualization of these orders is in Fig.~\ref{fig:generation_order}.

The experimental results are shown in Table~\ref{table:supp_generation_orders}. It indicates that a fully randomized order, the default choice of RandAR, performs the best. We conjecture that a fully randomized order best leverages the RandAR's capability of combining contexts from different locations of the images. Interestingly, the generation orders that encourage more divergent token locations at initial steps, \emph{i.e.}, hierarchical random and subsample, perform better than the other orders, potentially via covering a larger range of image contexts.

\subsection{Effects of parallel decoding}

\begin{figure}
    \centering
    \includegraphics[width=0.99\linewidth]{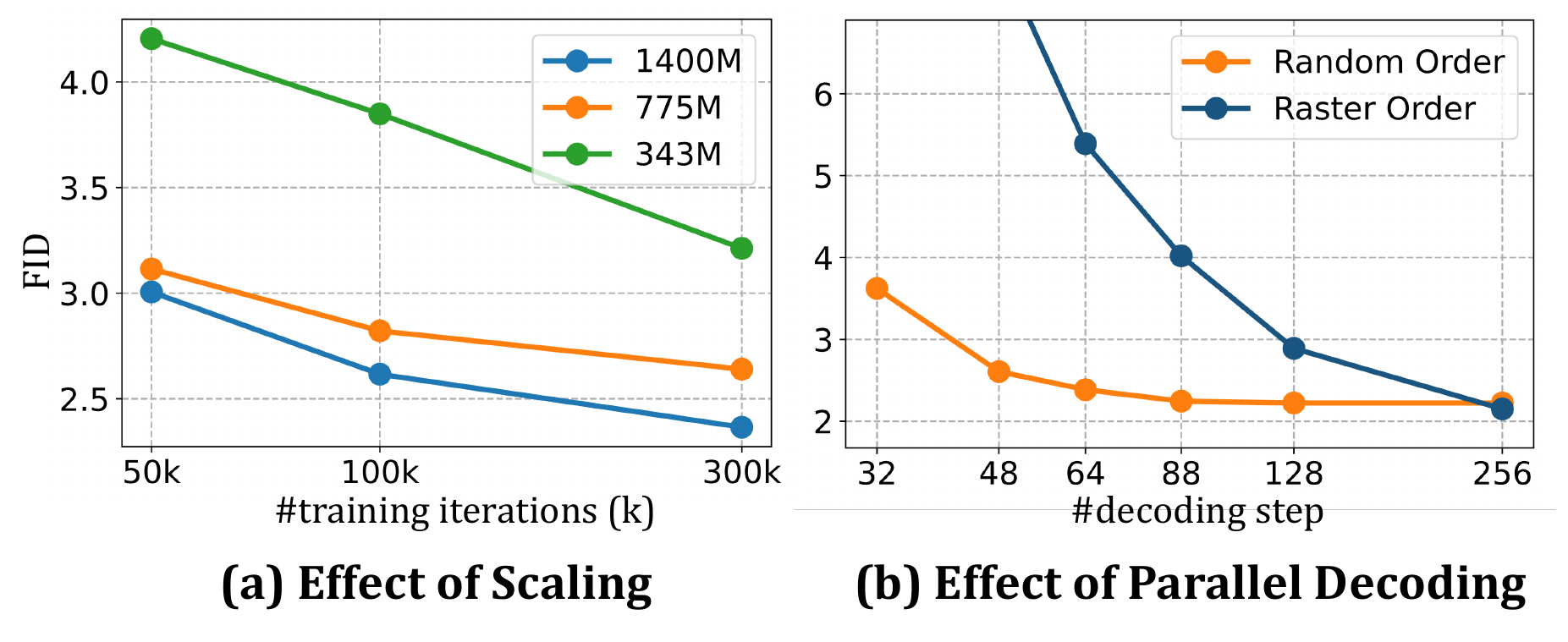}
    \vspace{-2mm}
    \caption{ \textbf{(a)} FID-50K over training iterations for RandAR in three different sizes. \textbf{(b)} 
    Effect of inference steps on FID-50K for RandAR and the raster-order counterpart (775M models).
 }
    \label{fig:scaling_pd}
    \vspace{-2mm}
\end{figure}

We apply parallel decoding to reduce inference steps and generation latency. To assess its impact on quality, we apply parallel decoding to both RandAR and a raster-order model trained under the same setup, reporting FID-50K. As shown in Fig.~\ref{fig:scaling_pd}\textcolor{cvprblue}{(b)}, the raster-order model experiences a significant performance drop, while RandAR maintains consistent quality up to 88 inference steps. This zero-shot ability comes from random order training.

To measure its improvement in generation speed, we assess latency with varied inference steps using PyTorch on an A100 GPU (40G VRAM). Specifically, we generate a batch of 64 images (equivalent to a batch size of 128 with Classifier-free guidance) at a 256x256 resolution. We also measure the latency of LLaMAGen~\cite{sun2024autoregressive} and MAR~\cite{li2024autoregressive} in the same way. For a fair comparison, the time spent on decoding latent image tokens to pixel space is omitted. As in Table~\ref{tab:latency}, RandAR using an 88-step schedule requires 2.9 $\times$ less steps and consequently lowers the latency by 2.5 $\times$.

\begin{table}
    \centering
\caption{Latency of generating 256$\times$256 images.  We test the latency on of our model with different decoding steps with Pytorch on A100 GPU (40G VRAM) and a batch size of 64 (128 with classifier-free guidance). 
Reducing inference steps with parallel decoding greatly lowers latency.
}\label{tab:latency}
\vspace{-2mm}
\resizebox{0.99\linewidth}{!}{
    \begin{tabular}{r|c@{\hspace{2mm}} c@{\hspace{2mm}} c@{\hspace{2mm}} c@{\hspace{4mm}} c@{\hspace{4mm}}}
    \toprule
      Method & Latency (sec.) & \#Params & \#Steps & Parallel Decoding &  KV-Cache  \\
         \midrule
    RandAR & 16.8   & 1.4B & 256 & Support & Support \\
    RandAR & 6.6  & 1.4B & 88  & Support & Support \\
    RandAR & 4.6  & 1.4B & 48  & Support & Support \\
    \midrule
    LLamaGen~\cite{sun2024autoregressive} &  15.9  & 1.4B & 256   & Noncompatible & Support \\
    \midrule
    MAR~\cite{li2024autoregressive} &  53.3  &  943M & 64  & Support & Noncompatible \\
    MAR~\cite{li2024autoregressive} &  220.0  &  943M & 256  & Support & Noncompatible \\
    \bottomrule
    \end{tabular}
    }
\end{table}

\subsection{Case Studies: Random v.s. Raster Order}
\label{sec:applications}

\subsubsection{Inpainting and Class-conditioned Editing}
\label{sec:exp_inpainting}

\begin{figure}
    \centering
    \includegraphics[width=0.99\linewidth]{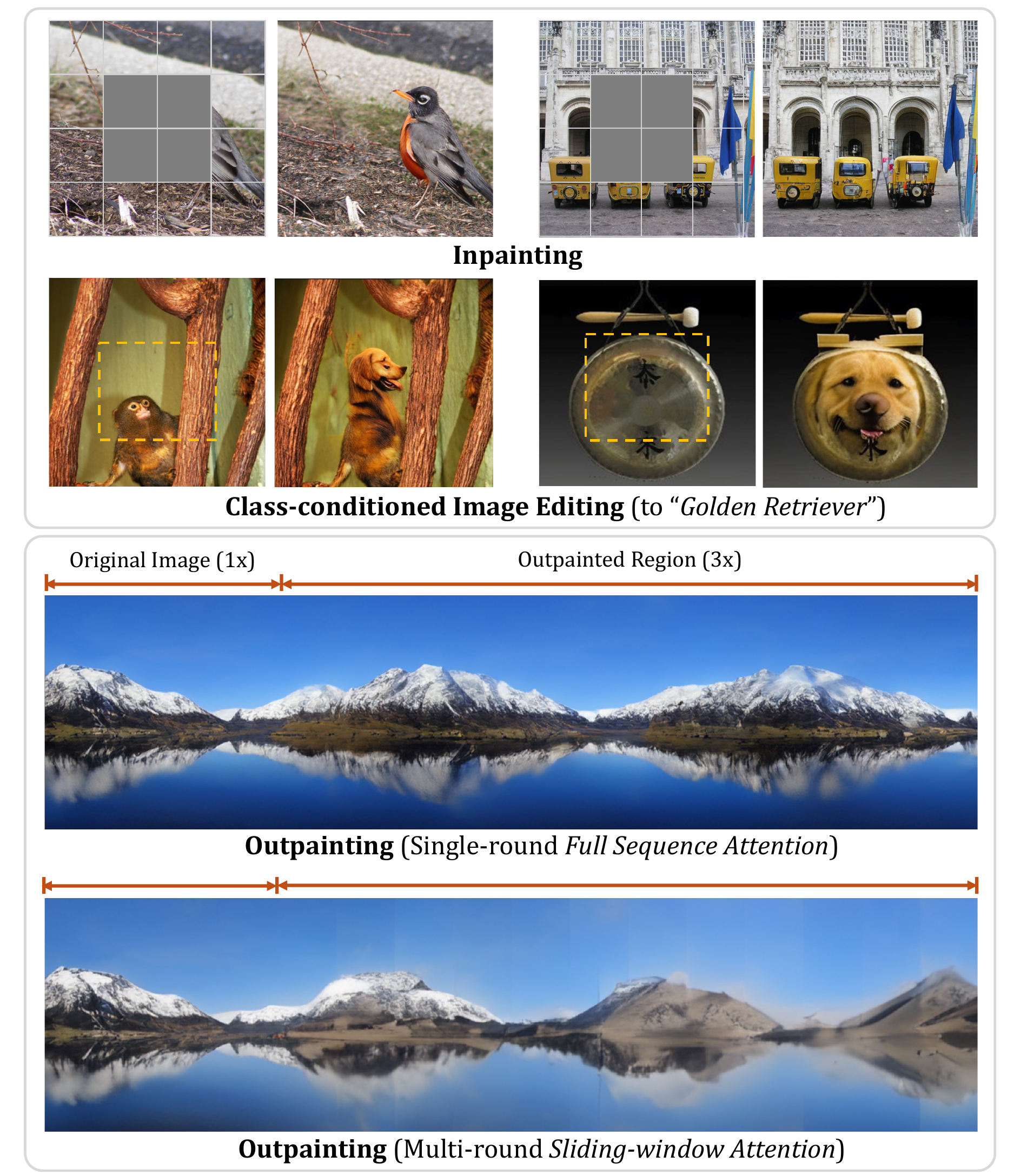}
    \vspace{-2mm}
    \caption{RandAR supports image manipulation tasks of inpainting and class-conditional image editing with a uni-directional transformer. Moreover, RandAR can use full casual attention for outpainting a 256$\times$256 image into a consistent 256$\times$1024.
    }
    \label{fig:inpainting}
\end{figure}

As described in Sec.~\ref{sec:inpainting}, RandAR can autoregressively fill blank patches in an image using all the visible tokens as context. Previously, only methods using bi-directional attention~\cite{chang2022maskgit, tian2024visual} demonstrated this capability. In Fig.~\ref{fig:inpainting}, we show zero-shot inpainting results from RandAR. 

\subsubsection{Outpainting} 
\label{sec:exp_outpainting}

As described in Sec.~\ref{sec:outpainting}, RandAR can extend an image beyond its boundaries directly using full causal sequence attention. In contrast, previous decoder-only visual AR models~\cite{esser2021taming} are limited to using partial context, relying only on image tokens to the left or above the target token, and typically employ an iterative sliding window attention approach. In Fig.~\ref{fig:inpainting}, we demonstrate outpainting by extending a 256$\times$256 image to 256$\times$1024 in the rightward direction. RandAR leverages \emph{full casual attention} for both source image context and newly generated tokens to maintain consistent patterns. By contrast, results from sliding window attention show a noticeable degradation in quality. More visualization examples are displayed in Sec.~\ref{sec:supp_outpainting}.

\begin{figure}
    \centering
    \includegraphics[width=0.99\linewidth]{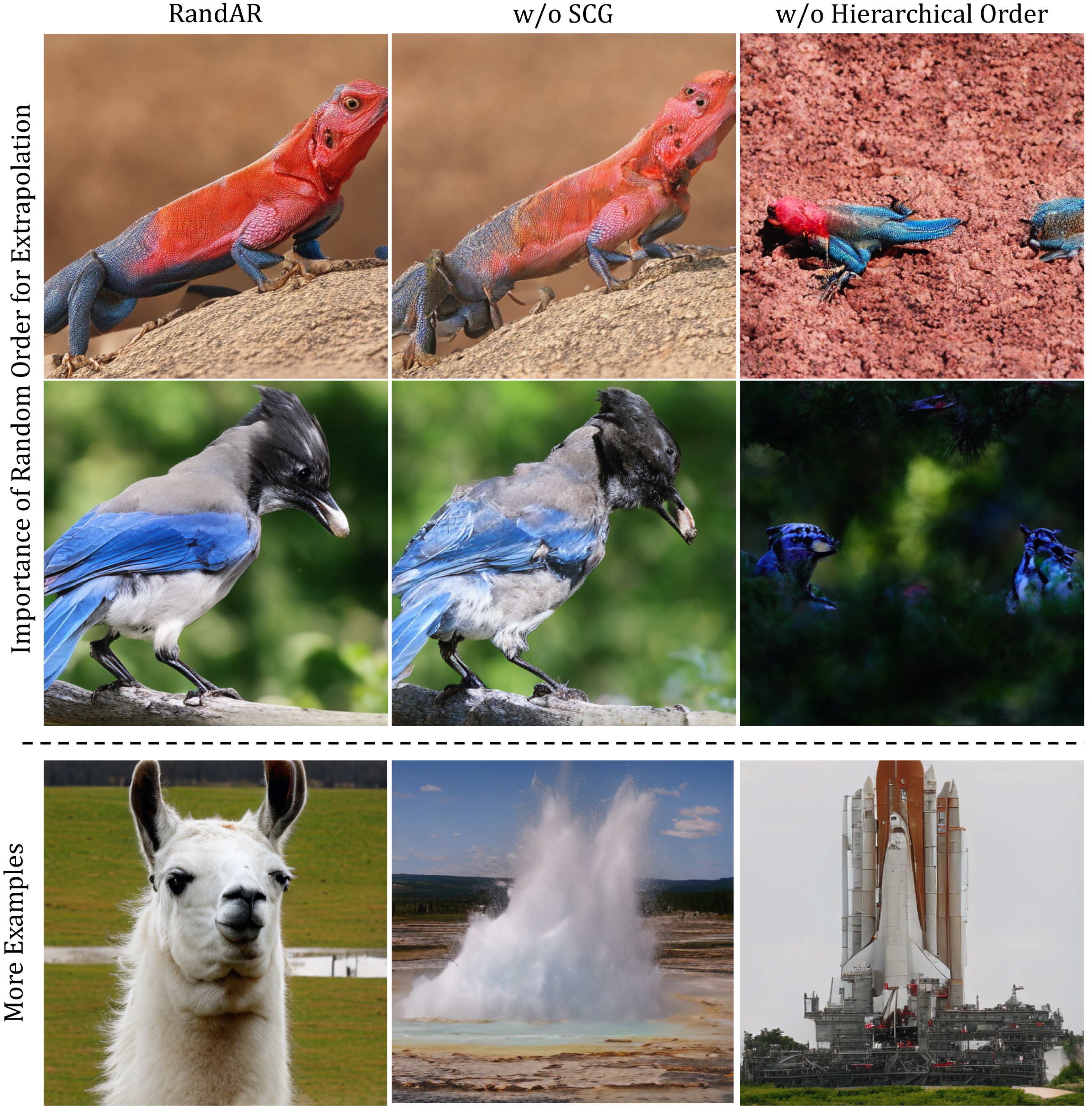}
    \vspace{-2mm}
    \caption{RandAR, trained on 256$\times$256 images, can generate 512$\times$512 images with finer details in zero-shot (Sec~\ref{sec:resolution}). This is achieved using a hierarchical decoding order, benefiting the unified layout,
    and SCG (Spatial Context Guidance) enhancing the visual quality by refining high-frequency details. 
    }
    \label{fig:resolution_extrapolation}
\end{figure}

\subsubsection{Resolution Extrapolation}
\label{sec:exp_resolution}

Fig.~\ref{fig:resolution_extrapolation} demonstrates RandAR’s zero-shot resolution extrapolation capability. Trained on 256$\times$256 images, RandAR can generate 512$\times$512 images with finer details. Unlike iterative outpainting~\cite{chang2022maskgit, zhang2023diffcollage, ding2023patched}, our approach does not aim to expand an image’s content but to enhance its detail with higher output resolution. The resolution extrapolation process uses a hierarchical order described in Sec.\ref{sec:resolution}. Images generated without this order, shown at the bottom of Fig.\ref{fig:resolution_extrapolation}, exhibit inconsistent patterns. The spatial contextual guidance (SCG) further enhances consistency in details, \emph{e.g.}, the two eyes of the chameleon. For more results, please refer to Sec.~\ref{sec:supp_extrapolation}. 
 
Zero-shot generalization to high-resolution images remains challenging, particularly in generating high-frequency patterns not prevalent in low-resolution training data. For instance, the space shuttle in Fig.~\ref{fig:resolution_extrapolation} struggles with intricate structures and boundaries. Addressing these challenges and supporting varied extrapolation ratios are areas for future exploration.

\subsubsection{Feature Encoding}
\label{sec:exp_bidirectional_encoding}

We compare the representations learned by random-order and raster-order generation models, both XL-sized models trained on ImageNet under the same setup. More detailed experiments are in Sec.~\ref{sec:supp_feature_encoding}.

\mypar{Local Representation.} We evaluate local representation using zero-shot semantic correspondence on the SPair71k benchmark~\cite{min2019spair}. SPair71k  contains nearly 71k paired images with sparse semantic-level correspondence annotated per pair. Following  DIFT~\cite{tang2023emergent}, We extract features for each image following Sec.~\ref{sec:bidirectional_encoding}, and detect correspondence using dot product between tokens from paired images.  We compute the “percentage of correct key points” (PCK), averaging the metrics per image or point.

As shown in Table~\ref{tab:feature_encoding}, by extracting features from the second round of tokens, RandAR effectively improves the correspondences. On the contrary, the raster order model experiences a significant drop and struggles to understand longer sequence lengths than training time. Although the random order model under-performs the raster one in the first round, possibly due to learning from a more challenging combination of orders, its features are finally better than a raster-order model with sufficient context. This shows that RandAR can directly generalize to bi-directional contexts without additional training. 

\mypar{Global Representation.} We average-pool the embeddings and perform linear probing, following MAE~\cite{he2022masked} on ImageNet~\cite{deng2009imagenet}. As shown in Table~\ref{tab:feature_encoding}, the global representation follows a similar trend as the local features: RandAR successfully generalizes to the bi-directional context in the second round of tokens, while the raster-order model fails to leverage additional context. This experiment further supports the idea that random-order unidirectional transformers can learn to model bi-directional contextual information.

\begin{table}[tb]
\centering
\caption{Random order training enables a \emph{uni-directional} decoder-only transformer to extract \emph{bi-directional} context by passing image token sequence twice through the model (2nd round), while raster-order models fail to extract bi-directional contextual features. }\label{tab:feature_encoding}
\vspace{-1mm}
\resizebox{0.99\linewidth}{!}{
\begin{tabular}{l|c@{\hspace{3mm}}c@{\hspace{3mm}}|c@{\hspace{3mm}}c@{\hspace{3mm}}}
\toprule
\multirow{2}{*}{Model} & \multicolumn{2}{c|}{Feature Correspondence (SPair71k)} & \multicolumn{2}{c}{Linear Probing (ImageNet)}  \\
& PCK (Per Image) $\uparrow$ & PCK (Per Point) $\uparrow$ & Top-1 Acc $\uparrow$ & Top-5 Acc $\uparrow$\\ 
\midrule
RasterAR & 24.5 & 28.6 & 62.6 & 83.9 \\
\rowcolor{gray!10}
w/ 2nd Round & 3.6 & 3.9 & 58.3 & 80.7 \\
\midrule
RandAR & 22.1 & 25.8 & 57.3  & 80.3 \\
\rowcolor{gray!10}
w/ 2nd Round & \textbf{31.3} & \textbf{36.4} & \textbf{63.1} & \textbf{84.2} \\
\bottomrule
\end{tabular}
}
\vspace{-3mm}
\end{table}

\section{Conclusions}
We introduce RandAR, a GPT-style causal decoder-only transformer that generates image tokens autoregressively in random orders. Our RandAR achieves this with specially designed position instruction tokens representing the location of next-token prediction. Despite the challenges of learning random order generation, RandAR achieves comparable performance with raster-order counterparts. Moreover, RandAR shows several new zero-shot applications for decoder-only models, including parallel decoding for 2.5$\times$ acceleration, inpainting, outpainting, resolution extrapolation, and feature extraction with bi-directional contexts. We hope RandAR inspires further exploration of unidirectional decoder-only models for visual tasks.

\mypar{Limitations and Future Works} Our RandAR investigates enabling decoder-only transformers to generate image tokens in random orders. Although it illustrates the advantages of combining bi-directional contexts from images, random-order generation so far achieves comparable performance compared with the raster-order counterparts, as learning from a much larger number of orders is significantly more challenging. Therefore, a meaningful future investigation would be improving the data efficiency of training a random-order model. 

In addition, we notice the trend of joint visual language generation with decoder-only transformers~\cite{wang2024emu3, zhou2024transfusion, team2024chameleon}, which uniformly follows a raster-order design. From this aspect, RandAR can be further scaled up from ImageNet pre-training to the image-text and image-video datasets.

\clearpage
\maketitlesupplementary

\renewcommand\thesection{\Alph{section}}
\renewcommand\thetable{\Alph{table}}
\renewcommand\thefigure{\Alph{figure}}
\renewcommand\thealgorithm{\Alph{algorithm}}
\renewcommand\theequation{\Alph{equation}}
\setcounter{section}{0}
\setcounter{table}{0}
\setcounter{figure}{0}
\setcounter{equation}{0}

In this supplementary material, we first provide the pseudo-code (Sec.~\ref{sec:supp_pseudo}) and additional implementation details (Sec.~\ref{sec:additional_details}). Then, we provide results at intermediate training steps (Sec.~\ref{sec:interm_steps_training}) and 384$\times$384 resolution generation performance (Sec.~\ref{sec:384_res}). We conduct a more in-depth analysis of bi-directional feature encoding (Sec.~\ref{sec:supp_feature_encoding}) and spatial contextual guidance (Sec.~\ref{sec:scg}). We finally present additional visualization of the generated images (Sec.~\ref{sec:generation_results}). 

\section{Pseudo-Code}\label{sec:supp_pseudo}

We have provided the Pytorch-style pseudo-code for our random-order training (Algorithm~\ref{alg:training}) and parallel decoding inference (Algorithm~\ref{alg:prallel_decoding}). We will release the code upon acceptance.
\begin{algorithm*}[tb]
\caption{RandAR Training Pytorch-style Pseudo-Code.}\label{alg:training}
\newcommand{\hlbox}[1]{%
  \fboxsep=1.2pt\hspace*{-\fboxsep}\colorbox{blue!10}{\detokenize{#1}}%
}
\lstset{style=mocov3}
\vspace{-3pt}
\begin{lstlisting}[
    language=python,
    escapechar=@,
    label=code:training]
    # Random order training. 
    # Input list: 
    # class_indices: [b, 1], b is batch size, dtype of torch.long; 
    # b, h, w: int, batch_size, height and width for latent space size
    # img_token_indices: [b, h * w], image token indices after the tokenizer
    # d: the hidden dimension of the model
    # head_dim: dimension of each attention head
    # model: the decoder-only transformer
    # Output: training loss

    # Step-1: Sample random orders
    seq_len = h * w
    raster_order_indices = torch.arange(seq_len).repeat(b, 1) # [b, seq_len]
    position_indices = random_permute(raster_order_indices)   # [b, seq_len]

    # Step-2: Prepare embeddings
    image_tokens = model.token_embeddings[image_token_indices] # [b, seq_len, d]
    image_tokens = torch.gather(image_tokens.unsqueeze(-1), dim=1, position_indices.unsqueeze(-1)) 
    cls_token = model.cls_embeddings[class_indices] # [b, d]

    # Random dropout
    image_tokens = random_dropout(image_tokens, p=0.1)
    cls_token = random_dropout(cls_token, p=0.1)
    
    # Step-3: Compute position instructions tokens
    # get 2D RoPE frequencies for each spatial location
    rope_freqs_cis = model.compute_rope_frequencies(b, h, w, base=10000) # [b, h, w, head_dim//2, 2]
    # flatten h, w to seq_len, arranging 2D RoPE frequencies in raster order
    rope_freqs_cis = rope_freqs_cis.flatten((1, 2)) # [b, seq_len, head_dim//2, 2]
    # get 2D rope frequencies in permuted random orders
    rope_freqs_cis = rope_freqs_cis[position_indices]
    
    # get position instruction tokens corresponding to tokens in random order
    pos_instruct_tokens = apply_2d_rope(model.shared_pos_embed, rope_freqs_cis) # [b, seq_len, d]

    # Step 4: Prepare Teacher Forcing Sequences
    x = torch.zeros(b, 1 + 2 * seq_len, d).to(image_tokens.device)
    x[:, 0] = cls_token
    x[:, 1::2] = pos_instruct_tokens
    x[:, 2::2] = image_tokens

    x_rope_freqs = torch.zeros(b, 1 + 2 * seq_len, head_dim // 2)
    x_rope_freqs[:, 0] = model.class_rope_freqs
    x_rope_freqs[:, 1::2] = rope_freqs_cis # rope for position instruction tokens
    x_rope_freqs[:, 2::2] = rope_freqs_cis # rope for image tokens

    # Step-5: Training with Next-token Prediction 
    pred_logits = model(x, x_rope_freqs) # [b, 1 + 2 * seq_len, vocab_size]
    
    # generated tokens from position instruction tokens
    pred_logits = pred_logits[:, 1::2] # [b, seq_len, vocab_size]

    # return back to raster order sequence for loss computation
    index_to_raster_order = torch.argsort(position_indices) # [bs, seq_len]
    raster_pred_logits = torch.gather(pred_logits, dim=1, index_to_raster_order.unsqueeze(-1))

    loss = cross_entropy(raster_pred_logits.view(-1, vocab_size), image_token_indices.view(-1))
    return loss
\end{lstlisting}\vspace{-5pt}
\end{algorithm*}
\begin{algorithm*}[tb]
\caption{RandAR Parallel Decoding Pytorch-style Pseudo-Code.}\label{alg:prallel_decoding}
\newcommand{\hlbox}[1]{%
  \fboxsep=1.2pt\hspace*{-\fboxsep}\colorbox{blue!10}{\detokenize{#1}}%
}
\lstset{style=mocov3}
\vspace{-3pt}
\begin{lstlisting}[
    language=python,
    escapechar=@,
    label=code:parallel_decoding]
    # Parallel decoding with cosine step size schedule. Classifier-free guidance is omitted for simplicity. 
    # Input list: 
    # class_indices: [b, 1], b is batch size, dtype of torch.long; 
    # b, h, w: int, batch_size, height and width for latent space size
    # d: the hidden dimension of the model
    # model, vq_vae: the decoder-only transformer and the Vector quantized VAE
    # Output: a batch of generated images

    # Step-1: Sample random orders
    seq_len = h * w
    raster_order_indices = torch.arange(seq_len).repeat(b, 1) # [b, seq_len]
    position_indices = random_permute(raster_order_indices)   # [b, seq_len]
    
    # Step-2: Compute position instructions tokens
    # get 2D RoPE frequencies for each spatial location
    rope_freqs_cis = model.compute_rope_frequencies(b, h, w, base=10000) # [b, h, w, head_dim//2, 2]
    # flatten h, w to seq_len, arranging 2D RoPE frequencies in raster order
    rope_freqs_cis = rope_freqs_cis.flatten((1, 2)) # [b, seq_len, head_dim//2, 2]
    # get 2D rope frequencies in permuted random orders
    rope_freqs_cis = rope_freqs_cis[position_indices]
    # get position instruction tokens corresponding to tokens in random order
    pos_instruct_tokens = apply_2d_rope(model.shared_pos_embed, rope_freqs_cis) # [b, seq_len, d]

    # Step-3: Init KV-caches & Class_embedding & PlaceHolder for generated tokens
    max_token_length = 1 + seq_len * 2 # class_embedding + position instruction tokens + image tokens
    model.setup_KVcache(max_token_length, batch_size=b)
    class_embed = model.class_embedding(class_indices) # [b, 1, d]
    generated_code_indices = torch.zeros((b, seq_len), dtype=torch.long) # [b, seq_len], placeholder
    num_generated = 0

    # Step-4: Prefill: prepare input of first decoding iteration
    step_size = 1 # number of decoding tokens for next iteration. starting at one-token-each-time
    x = torch.cat([class_embed, pos_instruct_embeddings[:, 0:1]], dim=1) # [b, 2, d]
    x_rope_freqs = torch.cat([model.class_rope_freqs, rope_freqs_cis[:, 0:1]], dim=1) # [b, 2, head_dim//2, 2]
    kvcache_write_indices = torch.arange(2)

    # Step-5: Start decoding loop. Using Parallel decoding with Cosine step-size schedule 
    while num_generated < seq_len:
        pred_logits = model(x, x_rope_freqs, kvcache_write_indices) # [b, num_cur_tokens, vocab_size]
        pred_logits = pred_logits[:, -step_size:] # [b, num_query_tokens, vocab_size]

        sampled_indices = sample(pred_logits, temperature=1.0, topk=-1) # [b, step_size] in torch.long
        generated_code_indices[:, num_generated:num_generated+step_size] = sampled_indices
        sampled_tokens = model.token_embedding(sampled_indices) # [b, step_size, d]

        # prepare input x, x_rope_freqs, kvcache_write_indices for next iterations. 
        step_size_next = CosineSchedule(num_generated, seq_len)
        # suppose the step size of last iteration is 2, the model decoded two image tokens, denoting as i1, i2.
        # denote the position instruction token for these two decoded tokens as p1, p2.
        # Then in last iteration, the input tensor x is [...., p1, p2]. 
        
        # suppose the step size for the next iteration remains as 2, 
        # with new position instruction tokens p3 and p4, 
        # then the input tensor x for the next iteration would be [i1, p2, i2, p3, p4].
        # Note We rewrite the KV-cache corresponding to [i1, p2, i2], 
        # so that the effective KV-cache follows the interleave format: [..., p1, i1, p2, i2, ...],  
        # consistent with training format. 
        
        # the number of input tokens for next iteration would be: 2 * step_size + step_size_next -1
        x = torch.zeros((b, 2 * step_size + step_size_next -1, d))
        x_rope_freqs = torch.zeros((b, 2 * step_size + step_size_next -1, head_dim // 2, 2))
        kvcache_write_indices = torch.arange(2 * step_size + step_size_next -1) + kvcache_write_indices[1-step_size]
        # using examples in above comments, fill in the [i1, p2, i2] part
        x[:, 0] = sampled_tokens[:, 1]
        x_rope_freqs[:, 0] = rope_freqs_cis[:, num_generated]
        for i in range(step_size - 1):
            x[:, 2 * i + 1] = pos_instruct_tokens[:, num_generated + i + 1]
            x[:, 2 * i + 2] = sampled_tokens[:, i + 1]
            x_rope_freqs[:, 2 * i + 1] = rope_freqs_cis[:, num_generated + i + 1]
            x_rope_freqs[:, 2 * i + 2] = rope_freqs_cis[:, num_generated + i + 1]

        num_generated += step_size
        step_size = step_size_next
        # using examples in above comments, fill in the  [p3, p4] part
        x[:, -step_size_next:] = pos_instruct_tokens[:, num_generated:num_generated + step_size_next]
        x_rope_freqs[:, -step_size_next:] = rope_freqs_cis[:, num_generated:num_generated + step_size_next]

    # Step-6, decode generated tokens to images
    index_to_raster_order = torch.argsort(position_indices) # [bs, seq_len, 1]
    generated_code_indices = torch.gather(generated_code_indices.unsqueeze(-1), dim=1, index_to_raster_order)
    img = vq_vae.decode(generated_code_indices)
\end{lstlisting}\vspace{-5pt}
\end{algorithm*}

\section{Additional Implementation Details}
\label{sec:additional_details}

For image preprocessing, we follow the approach in LlamaGen~\cite{sun2024autoregressive}. Specifically, for 256$\times$256 experiments, we resize the image’s shorter edge to  $256 \times 1.1$ and apply a center square crop with the same size, and then perform ten-crop augmentation.

All the experiments in Table~\ref{table:main_results}, including raster-order counterparts, use classifier-free guidance (CFG) with a linear schedule for sampling. The optimal CFG weight is determined through a sweep with a step size of 0.1 across all methods. We use a temperature of 1.0 without applying top-$k$ filtering.

\mypar{Model Configurations.} We have provided several sizes of the RandAR model following LLaMAGen~\cite{sun2024autoregressive}, which are plain decoder-only transformers. The detailed architectures are in Table~\ref{tab:model_scale}.

\begin{table}[h]
\centering
\caption{Model configurations of RandAR.}\label{tab:model_scale}
\vspace{-3mm}
\resizebox{0.99\linewidth}{!}{
\begin{tabular}{l|cccc}
\toprule
\textbf{Model} & \textbf{Parameters} & \textbf{Layers} & \textbf{Hidden Dim} & \textbf{Attn Heads} \\
\midrule
RandAR-L & 343M & 24 & 1024 & 16 \\
RandAR-XL & 775M & 36 & 1280 & 20 \\
RandAR-XXL & 1.4B & 48 & 1536 & 24 \\
\bottomrule
\end{tabular}
}
\vspace{-2mm}
\end{table}

\begin{figure}[h]
    \centering
    \includegraphics[width=0.99\linewidth]{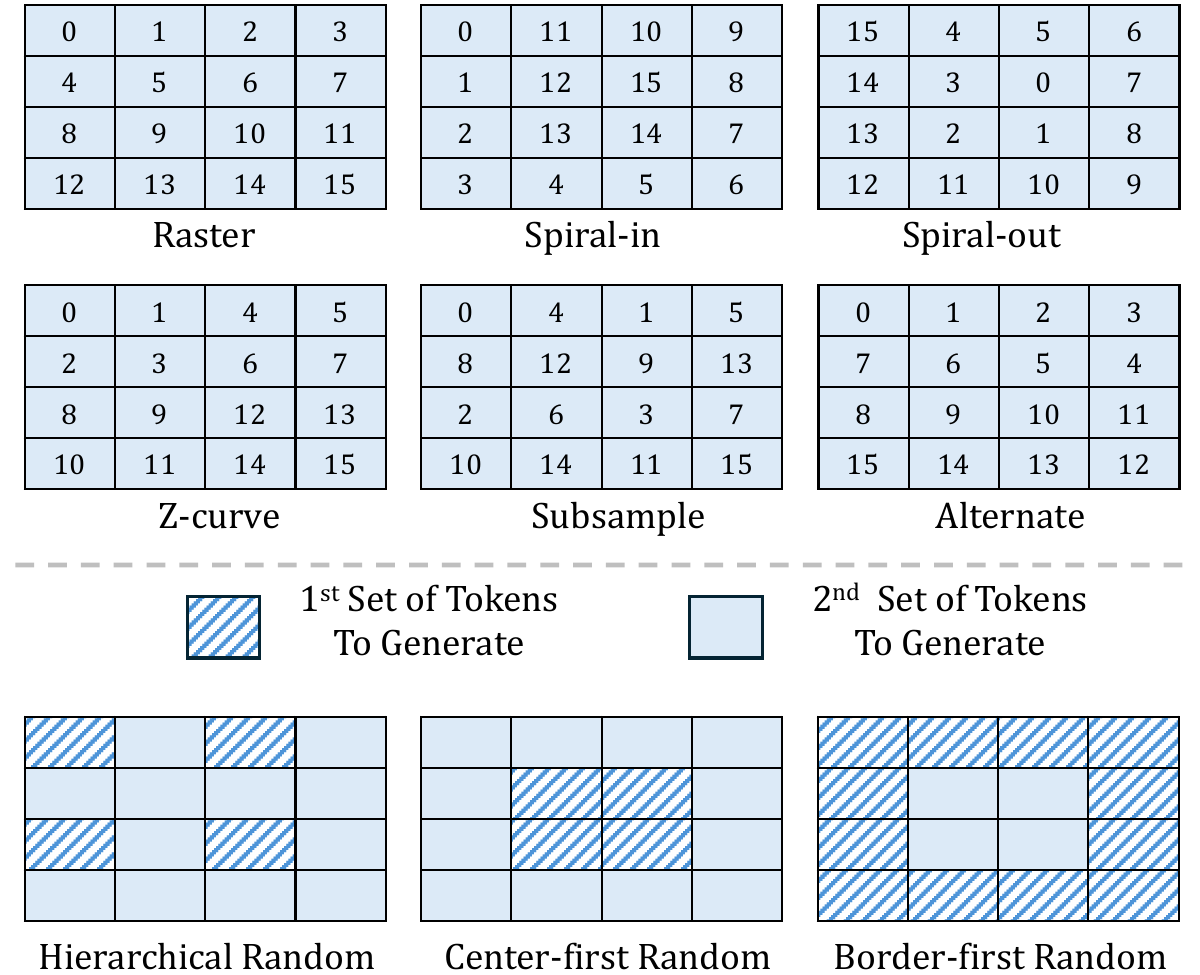}
    \vspace{-2mm}
    \caption{Illustration of different generation orders. RandAR, by default, uses a fully randomized order for inference. We investigate the fixed generation orders proposed in VQGAN~\cite{esser2021taming} (top) and partially random orders guided by the priors of hierarchy, center-first, and border-first (bottom).}
    \label{fig:generation_order}
\end{figure}

\mypar{Generation Orders.} In Table~\ref{table:supp_generation_orders}, we analyze RandAR with different generation orders. These orders are visualized in Fig.~\ref{fig:generation_order} using 4$\times$4 grids.

\section{Performance at Middle Training Steps}\label{sec:interm_steps_training}

We provide the additional results of our RandAR models at intermediate training steps in Table~\ref{table:supp_interm_training}. All the results are evaluated with 88 steps of generation with parallel decoding described in Sec.~\ref{sec:parallel_decoding}.
\begin{table}[t]
\centering
\caption{Generation Results for Intermediate Training Steps. With a batch size of 1024, 300 epochs of full RandAR training equals 360 iterations.}\vspace{-3mm}
\label{table:supp_interm_training}
\resizebox{0.99\linewidth}{!}{
\begin{tabular}{l|c|c|c|c|c}
\toprule
\textbf{Model} & \textbf{Iters} & \textbf{FID↓} & \textbf{IS↑} & \textbf{Precision↑} & \textbf{Recall↑} \\
\midrule
\multirow{3}{*}{RandAR-L} & 50k & 4.21 & 224.2 & 0.83 & 0.49 \\
 & 100k & 3.85 & 251.4 & 0.82 & 0.52 \\
 & 300k & 3.21 & 259.7 & 0.80 & 0.55 \\
\midrule
\multirow{3}{*}{RandAR-XL} & 50k & 3.11 & 271.1 & 0.81 & 0.53 \\
 & 100k & 2.82 & 293.6 & 0.81 & 0.56 \\
 & 300k & 2.66 & 296.3 & 0.80 & 0.57 \\
\midrule
\multirow{3}{*}{RandAR-XXL} & 50k & 3.01 & 277.4 & 0.79 & 0.57 \\
 & 100k & 2.61 & 296.5 & 0.79 & 0.57 \\
 & 300k & 2.37 & 309.5 & 0.79 & 0.60 \\
\bottomrule
\end{tabular}
}
\end{table}

\section{RandAR on ImageNet at 384 Resolution}
\label{sec:384_res}

We report results on ImageNet at 384×384 resolution using the same tokenizer, which produces 24×24 tokens per image, corresponding to 576! random permutations. Our XL-sized model, with 775M parameters, is trained using the same setup. With 180 sampling steps, the model achieves an FID of 2.32 and an Inception Score of 323. Using 144 sampling steps, it achieves an FID of 2.35 and an Inception Score of 322. The FID is slightly higher than that of the 256×256 model, consistent with observations in LlamaGen~\cite{sun2024autoregressive} that models smaller than 1B parameters perform slightly worse at 384×384 resolution.

\vspace{-2mm}
\section{Additional Results on Feature Encoding}
\label{sec:supp_feature_encoding}
\vspace{-2mm}

We have conducted feature encoding experiments in Sec.~\ref{sec:exp_bidirectional_encoding} and Table~\ref{tab:feature_encoding}, suggesting that decoder-only transformers learned in random generation orders can generalize to extracting features from bi-directional contexts, while raster-order models cannot. In this section, we provide additional ablation studies and discussions.

\mypar{Comparison with VQ Tokenizer.} Here we demonstrate that our autoregressive transformer learns better representation than its VQ tokenizer, which provides the input token indices to our RandAR transformers. Specifically, we conduct the feature correspondence experiment on SPari71k~\cite{min2019spair} with the DIFT~\cite{tang2023emergent} framework, only replacing the feature extractor with the encoder from the VQ Tokenizer. As shown in Table~\ref{tab:supp_feature_encoding}, VQ Tokenizer performs significantly worse than our RandAR.

\mypar{Transformer Layers in RandAR.} As noticed by previous work~\cite{zou2024segment, pang2024frozen}, varied layers from a decoder-only language model can have significantly different abilities for visual feature encoding. RandAR has a similar case since the earlier layers might concentrate on low-level patterns while later layers are primarily used to map the features into the token space. For our 775M model, which has 36 layers, we analyze the performance difference for 12-th, 24-th, and 36-th layers. As shown in Table~\ref{tab:supp_feature_encoding}, the 24-th layer performs the best for both random and raster order models; thus, it is used for our comparison in Table~\ref{tab:feature_encoding} in the main paper. 

\begin{table}
\centering
\caption{Ablation Studies for Finding Feature Correspondences on SPair71k~\cite{min2019spair}. We search for the best decoder-only transformer learning for feature encoding (24-th), and then show that the features from our RandAR transformer are better than those from the VQ image tokenizer.}\label{tab:supp_feature_encoding}
\vspace{-3mm}
\resizebox{0.99\linewidth}{!}{
\begin{tabular}{l|c|c@{\hspace{3mm}}c@{\hspace{3mm}}}
\toprule
\multirow{2}{*}{Model} & \multirow{2}{*}{Layer} & \multicolumn{2}{c}{Feature Correspondence (SPair71k)} \\
& & PCK (Per Image) $\uparrow$ & PCK (Per Point) $\uparrow$ \\ 
\midrule
RasterAR & \multirow{4}{*}{24-th} & 24.5 & 28.6 \\
w/ 2nd Round &  & 3.6 & 3.9 \\
\cmidrule{1-1} \cmidrule{3-4} 
RandAR & & 22.1 & 25.8 \\
\rowcolor{gray!10}
w/ 2nd Round & & \textbf{31.3} & \textbf{36.4}  \\
\midrule
VQ Tokenizer & - & 5.6 & 6.0 \\
\midrule
RasterAR & \multirow{4}{*}{12-th} & 10.7 & 12.4 \\
RasterAR w/ 2nd Round & & 3.5 & 3.7 \\
\cmidrule{1-1} \cmidrule{3-4}
RandAR w/ 2nd Round & & 16.3 & 19.0 \\ 
\rowcolor{gray!10}
RandAR  &  & 11.0 & 12.6 \\
\midrule
RasterAR & \multirow{4}{*}{36-th} & 11.1 & 12.5\\
RasterAR w/ 2nd Round & &  3.5 & 3.7 \\
\cmidrule{1-1} \cmidrule{3-4} 
RandAR &  & 2.4 & 2.5 \\
\rowcolor{gray!10}
RandAR w/ 2nd Round & & 10.3 & 11.2 \\
\bottomrule
\end{tabular}
}
\vspace{-3mm}
\end{table}

\vspace{-2mm}
\section{Spatial Contextual Guidance}
\label{sec:scg}
\vspace{-2mm}

In Sec.~\ref{sec:resolution}, we introduce a new type of guidance called ``\emph{Spatial Contextual Guidance}'' (SCG) inspired by the classifier-free guidance (CFG). This section describes SCG in detail and analyzes its benefits.

\subsection{Formulation of Spatial Contextual Guidance}

The motivation of SCG is to enable better consistency in high-frequency details, as shown in Fig.~\ref{fig:resolution_extrapolation}. Inspired by CFG, SCG guides the generation by calculating the difference between the two sampling results with \emph{all the previous tokens} as context and \emph{part of the previous tokens as context}. Denoting the RanAR network as $e_{\theta}(\cdot)$, the spatial contextual guidance is:
\begin{equation}
    \tilde{e}_{\theta}(\mathbf{x}_{1:n}, c) = e_{\theta}(\mathbf{x}^{\phi}_{1:n}, c) + w_{\mathrm{scg}}(e_{\theta}(\mathbf{x}_{1:n}, c) - (e_{\theta}(\mathbf{x}^{\phi}_{1:n}, c)),
\end{equation}
where $c$ is the class conditioning, $\mathbf{x}_{1:n}$ is the set of tokens generated in previous steps, and $\mathbf{x}^{\phi}_{1:n}$ is the set of tokens with a random dropout. With such guidance, the final generated result $\tilde{e}_{\theta}(\mathbf{x}_{1:n}, c)$ has better consistency with the tokens dropped out from $\mathbf{x}_{1:n}$.

When combining SCG with the conventional CFG, we follow InstructPix2Pix~\cite{brooks2023instructpix2pix} and Liu~\emph{et al.}~\cite{liu2022compositional} to compose two guidances together:
\begin{equation}
\label{eqn:scg}
\begin{aligned}
    \widetilde{e}_{\theta}(\mathbf{x}_{1:n}, c) = & e_{\theta}(\mathbf{x}_{1:n}^{\phi}, c^{\phi}) + \\ & w_{\mathrm{scg}} (e_{\theta}(\mathbf{x}_{1:n}, c^{\phi}) - e_{\theta}(\mathbf{x}_{1:n}^{\phi}, c^{\phi})) + \\
    & w_{\mathrm{cfg}} (e_{\theta}(\mathbf{x}_{1:n}, c) - e_{\theta}(\mathbf{x}_{1:n}, c^{\phi})),
\end{aligned}
\end{equation}
where $w_{\mathrm{scg}}=1$ will make the above guidance equivalent to conventional CFG.

SCG is supported by the training since the image tokens experience a random dropout of 10\%, following the standard practice in LLaMAGen~\cite{sun2024autoregressive}. During the inference time, we randomly dropout a token to all zeros by the probability of 25\% to create $\mathbf{x}_{1:n}^{\phi}$. In this way, the generation process is still fully autoregressive and compatible with KV-cache.

\vspace{-1mm}
\subsection{Evaluation of Spatial Contextual Guidance}
\vspace{-1mm}

SCG can improve the visual quality of regular resolution generation and resolution extrapolation.

\begin{figure}
    \centering
    \includegraphics[width=0.99\linewidth]{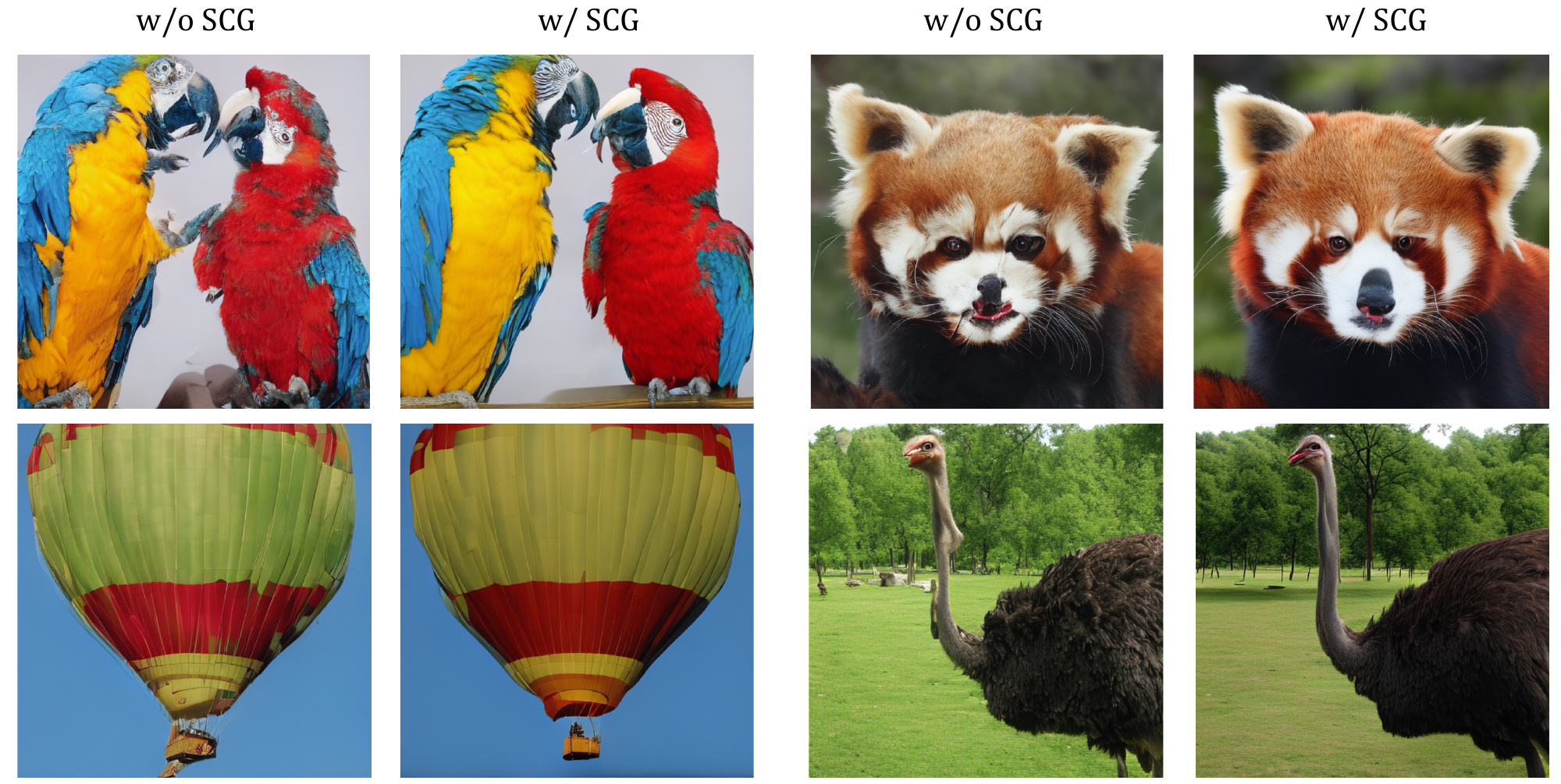}
    \vspace{-3mm}
    \caption{Using spatial contextual guidance (SCG) significantly improves the visual quality for zero-shot resolution extrapolation, especially the high-frequency details. The images are all 512$\times$512, and the ``w/ SCG'' samples are from $w_{\mathrm{scg}}=2.5$. (Zoom in for high-resolution details.)}
    \label{fig:supp_scg}
\end{figure}

\mypar{Resolution Extrapolation.} As shown in Fig.~\ref{fig:supp_scg}, using SCG enhances the high-frequency details of the images when generating 512$\times$512 images directly from our 775M RandAR trained from 256$\times$256. Numerous extraneous parts of the objects and uneven patterns are removed.

\mypar{Regular 256$\times$256 Generation.} Although SCG is primarily proposed for the challenging zero-shot resolution extrapolation, its effects are also reflected in regular 256$\times$256 generation. Our observation is also validated by quantitative evaluation. As in Table~\ref{table:supp_scg_256x256} evaluating the 775M RandAR model, SCG of $w_{\mathrm{scg}}=1.2$ can improve sFID, which emphasizes more low-level details, at a marginal drop of FID. 

\begin{table}[t]
\centering
\vspace{-3mm}
\caption{Ablation Study of Spatial Contextual Guidance (SCG). SCG can improve the visual quality for random-order AR models, where it decreases sFID by a large margin with a minor drop in FID.  
Although SCG is only designed for resolution extrapolation, it implicitly reflects the advantage of RandAR in combining bi-directional context.}\vspace{-3mm}
\label{table:supp_scg_256x256}
\resizebox{0.99\linewidth}{!}{
\begin{tabular}{l|c|c|c|c|c|c}
\toprule
\textbf{Model} & \textbf{SCG} & \textbf{FID↓} & \textbf{sFID↓} & \textbf{IS↑} & \textbf{Precision↑} & \textbf{Recall↑}\\
\midrule
\multirow{2}{*}{RandAR} & \xmark & 2.25 & 6.13 & 317.8 & 0.80 & 0.60 \\
 & \cmark & 2.34 & 5.85 & 303.8 & 0.80 & 0.60 \\
\bottomrule
\end{tabular}
}
\vspace{-3mm}
\end{table}

\vspace{-1mm}
\section{Additional Generation Results}\label{sec:generation_results}
\vspace{-1mm}
\subsection{Outpainting}\label{sec:supp_outpainting}

\begin{figure}[tb]
    \centering
    \includegraphics[width=0.99\linewidth]{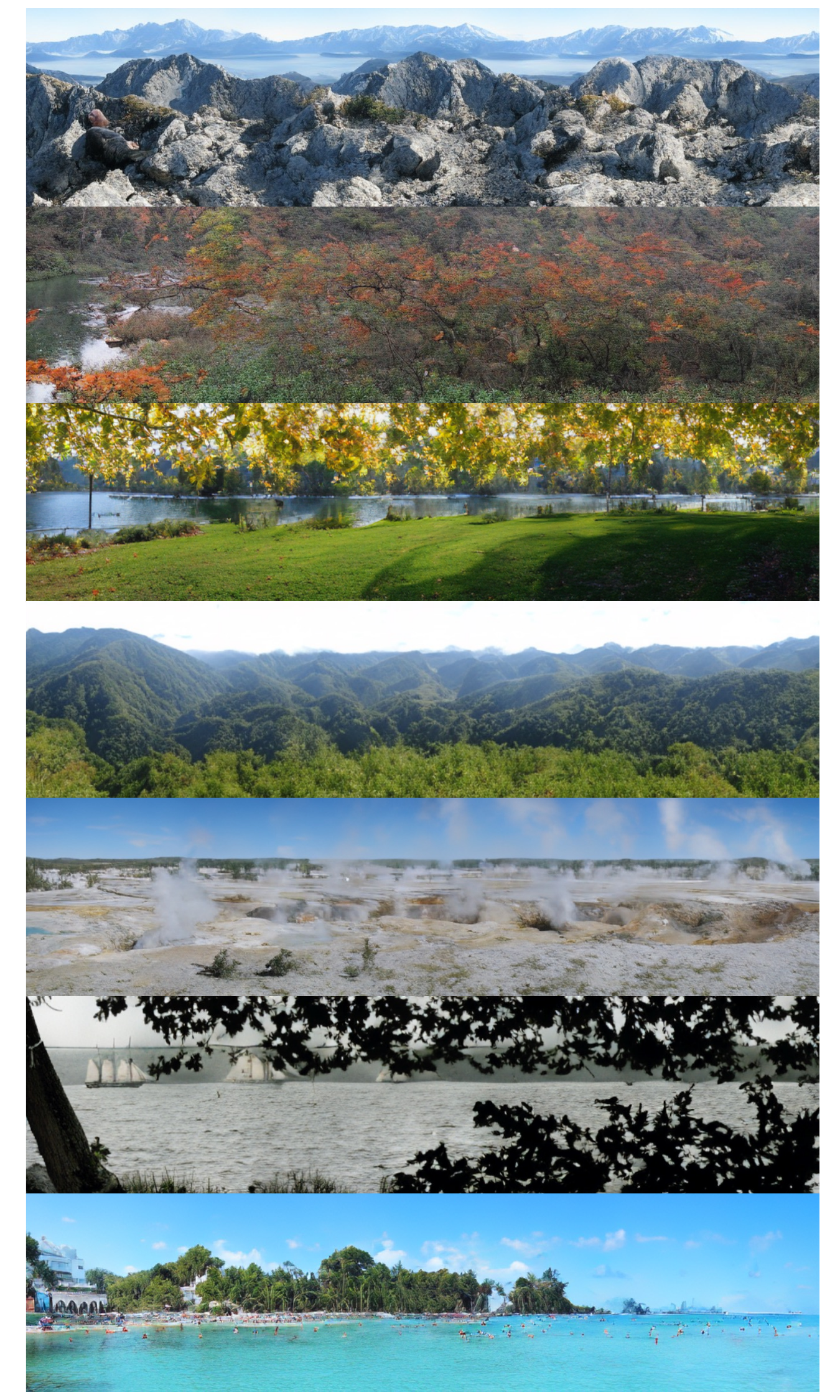}\vspace{-3mm}
    \caption{4$\times$ Outpainting results using 256$\times$256 RandAR to generate 256$\times$1024 images. Full sequence attention is used.}
    \label{fig:outpainting_full_attn}
\end{figure}

We provide additional visualizations of the outpainting results in Fig.~\ref{fig:outpainting_full_attn}.

\subsection{Regular Image Generation}

\begin{figure}[tb]
    \centering
    \includegraphics[width=0.99\linewidth]{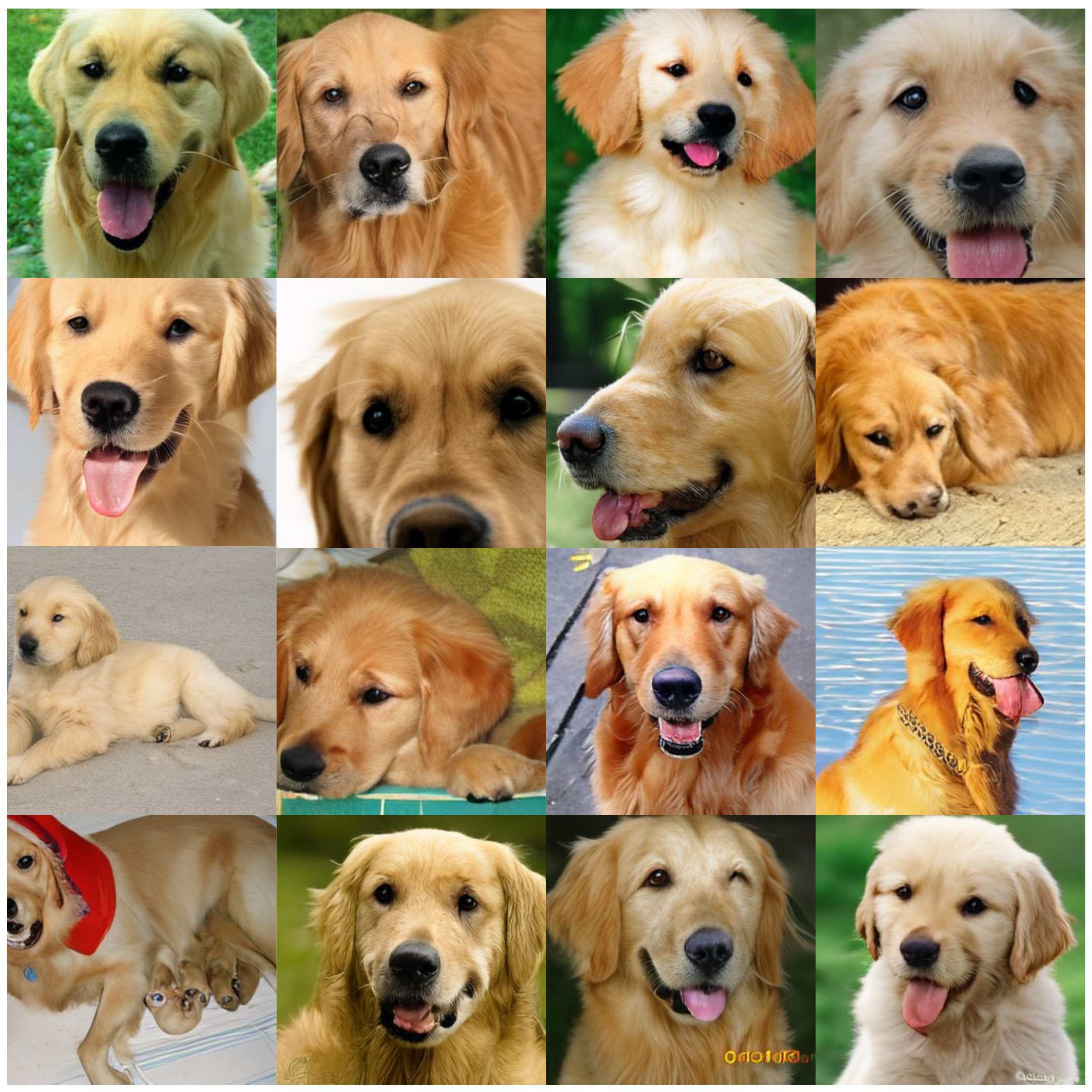}
    \caption{Uncurated generation results (256$\times$256). $w_{\mathrm{cfg}}=4.0$. Golden retriever (ImageNet class 207).}
    \label{fig:uncurated_207}
\end{figure}

\begin{figure}
    \centering
    \includegraphics[width=0.99\linewidth]{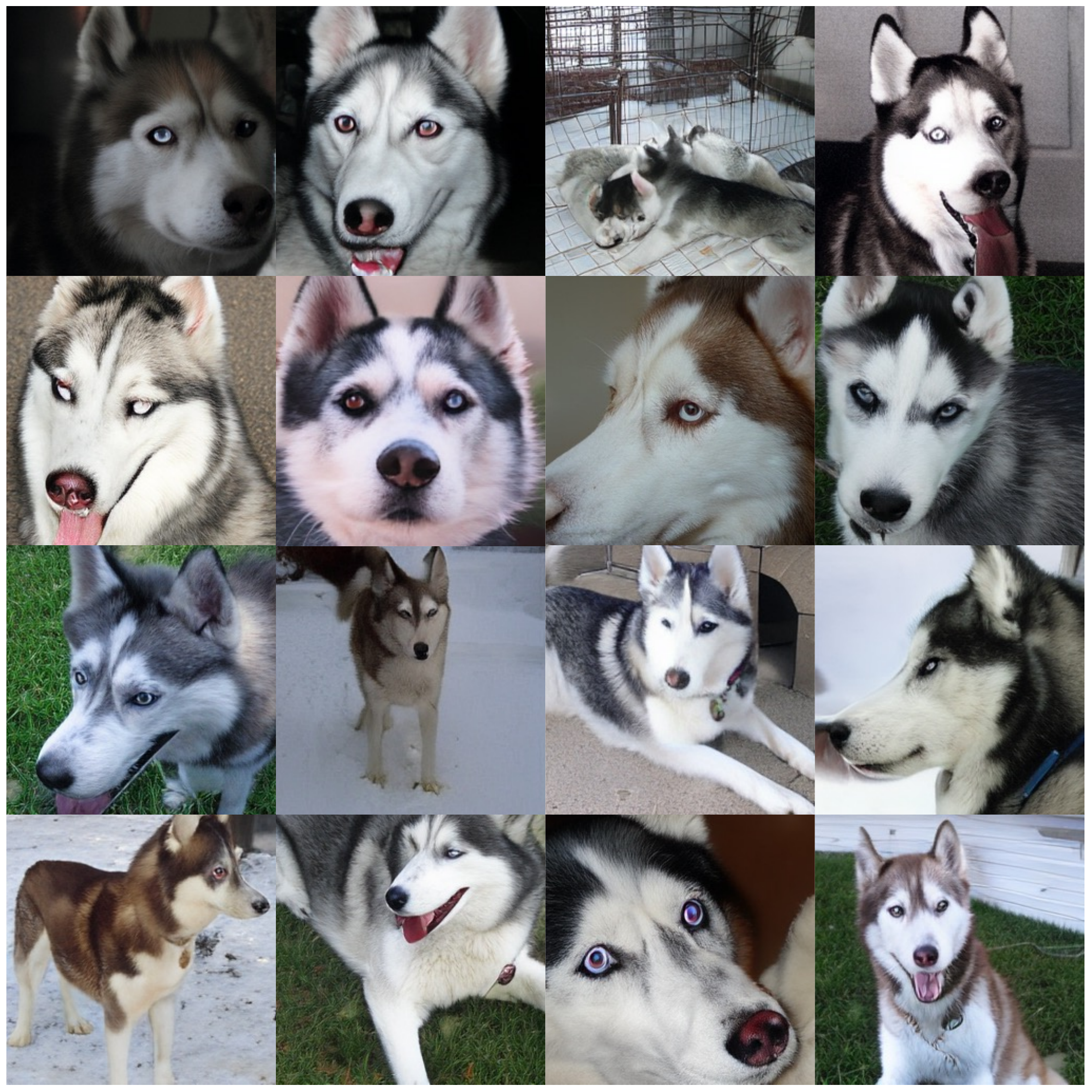}
    \caption{Uncurated generation results (256$\times$256). $w_{\mathrm{cfg}}=4.0$. Husky (ImageNet class 250).}
    \label{fig:uncurated_250}
\end{figure}

\begin{figure}
    \centering
    \includegraphics[width=0.99\linewidth]{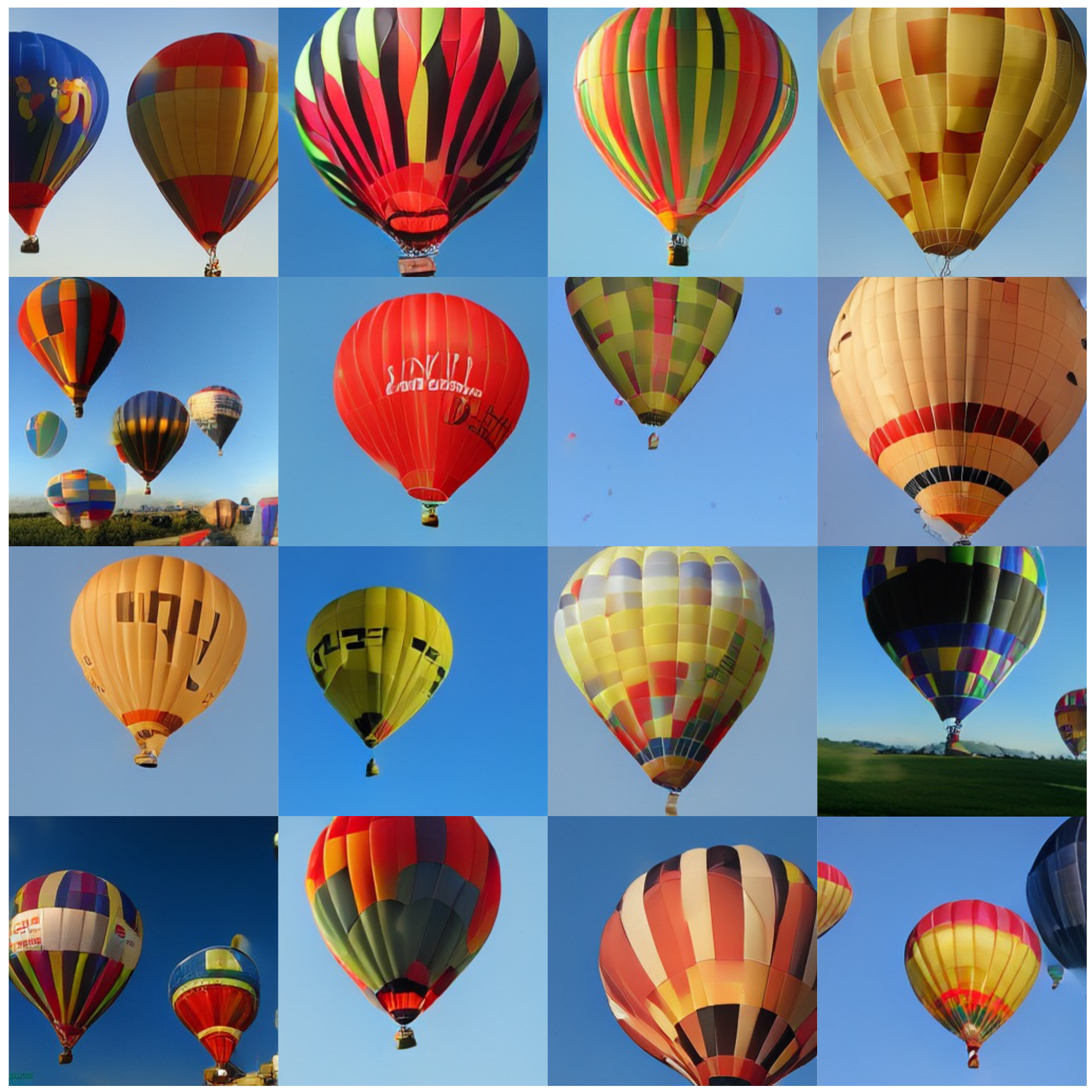}
    \caption{Uncurated generation results (256$\times$256). $w_{\mathrm{cfg}}=4.0$. Balloon (ImageNet class 417).}
    \label{fig:uncurated_417}
\end{figure}

\begin{figure}
    \centering
    \includegraphics[width=0.99\linewidth]{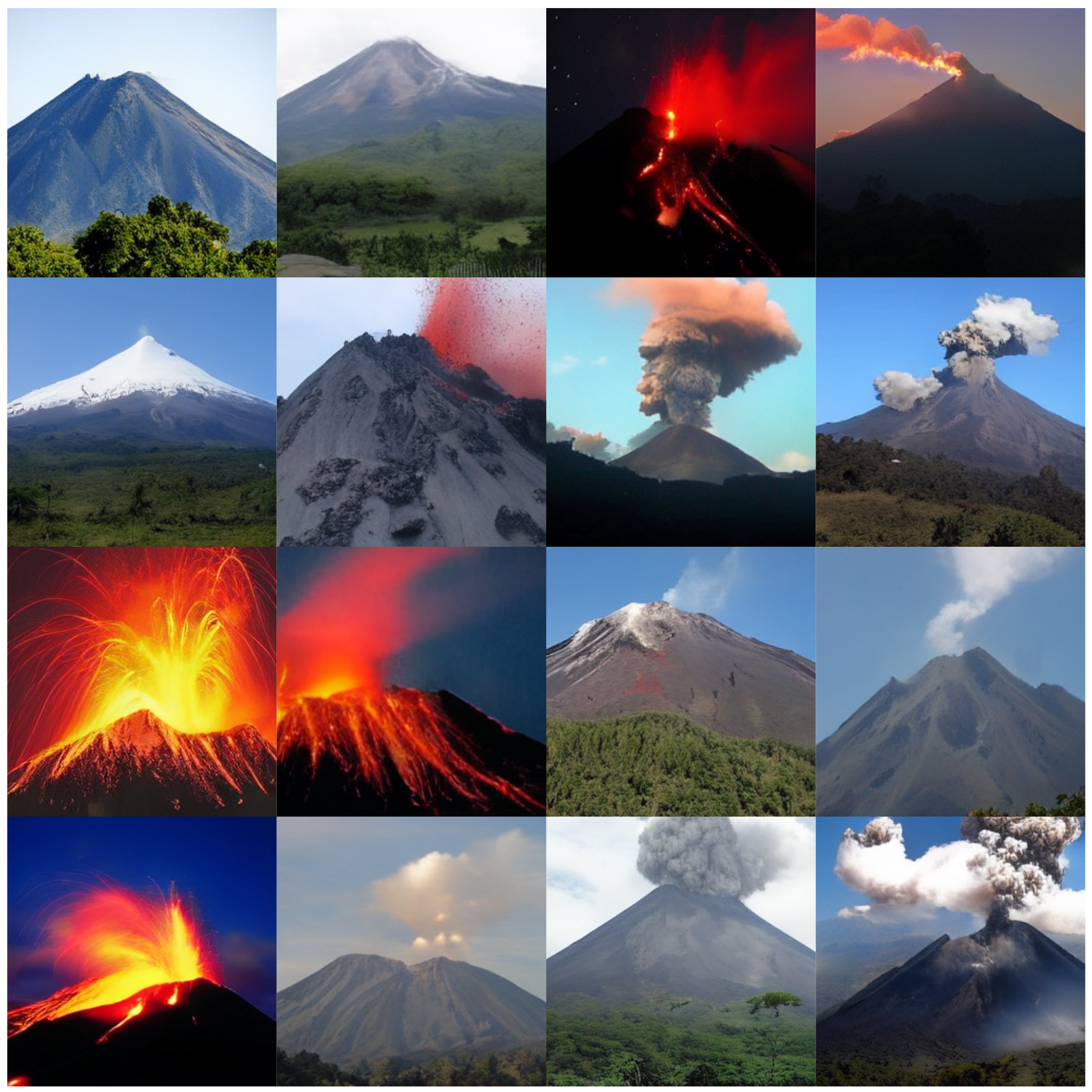}
    \caption{Uncurated generation results (256$\times$256). $w_{\mathrm{cfg}}=4.0$. Volcano (ImageNet class 980).}
    \label{fig:uncurated_980}
\end{figure}

\begin{figure}
    \centering
    \includegraphics[width=0.99\linewidth]{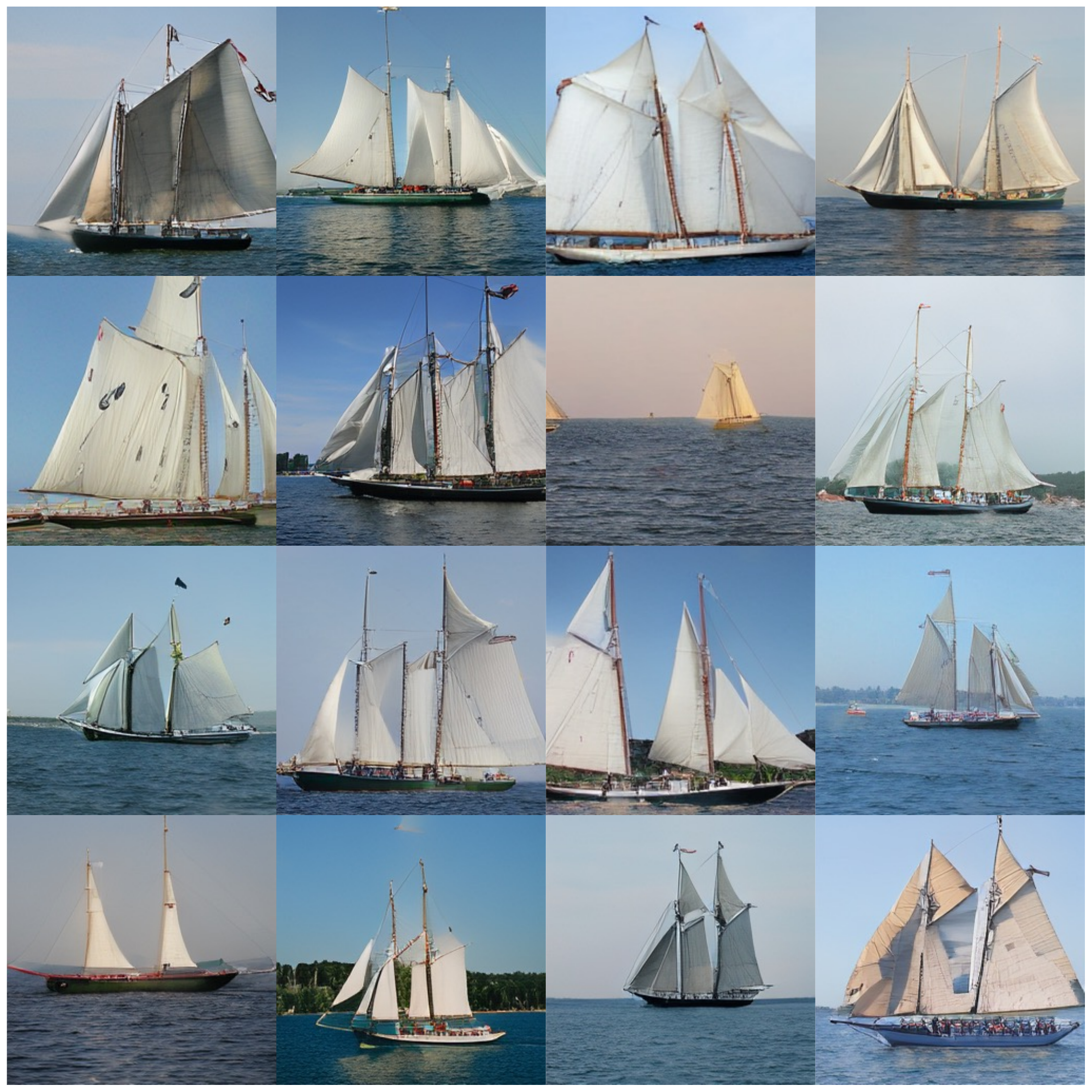}
    \caption{Uncurated Generation Results (256$\times$256). $w_{\mathrm{cfg}}=4.0$. Schooner (ImageNet class 780).}
    \label{fig:uncurated_780}
\end{figure}

\begin{figure}
    \centering
    \includegraphics[width=0.99\linewidth]{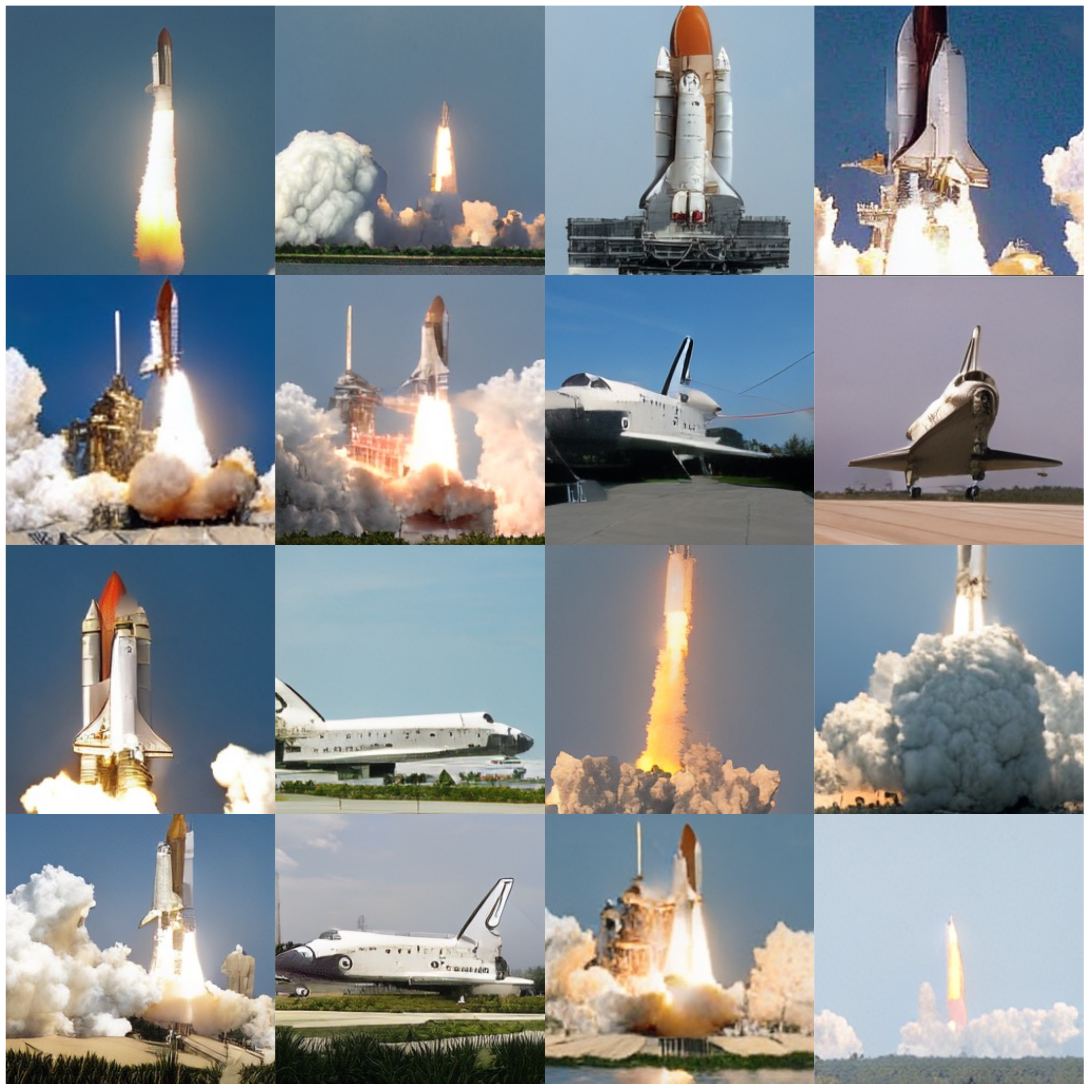}
    \caption{Uncurated Generation Results (256$\times$256). $w_{\mathrm{cfg}}=4.0$. Space shuttle (ImageNet class 812).}
    \label{fig:uncurated_812}
\end{figure}

We demonstrate the uncurated 256$\times$256 images generated from our 775M RandAR-XL. They are displayed from Fig.~\ref{fig:uncurated_207} to Fig.~\ref{fig:uncurated_812}. 

\subsection{Resolution Extrapolation}\label{sec:supp_extrapolation}

\begin{figure}
\centering\includegraphics[width=0.99\linewidth]{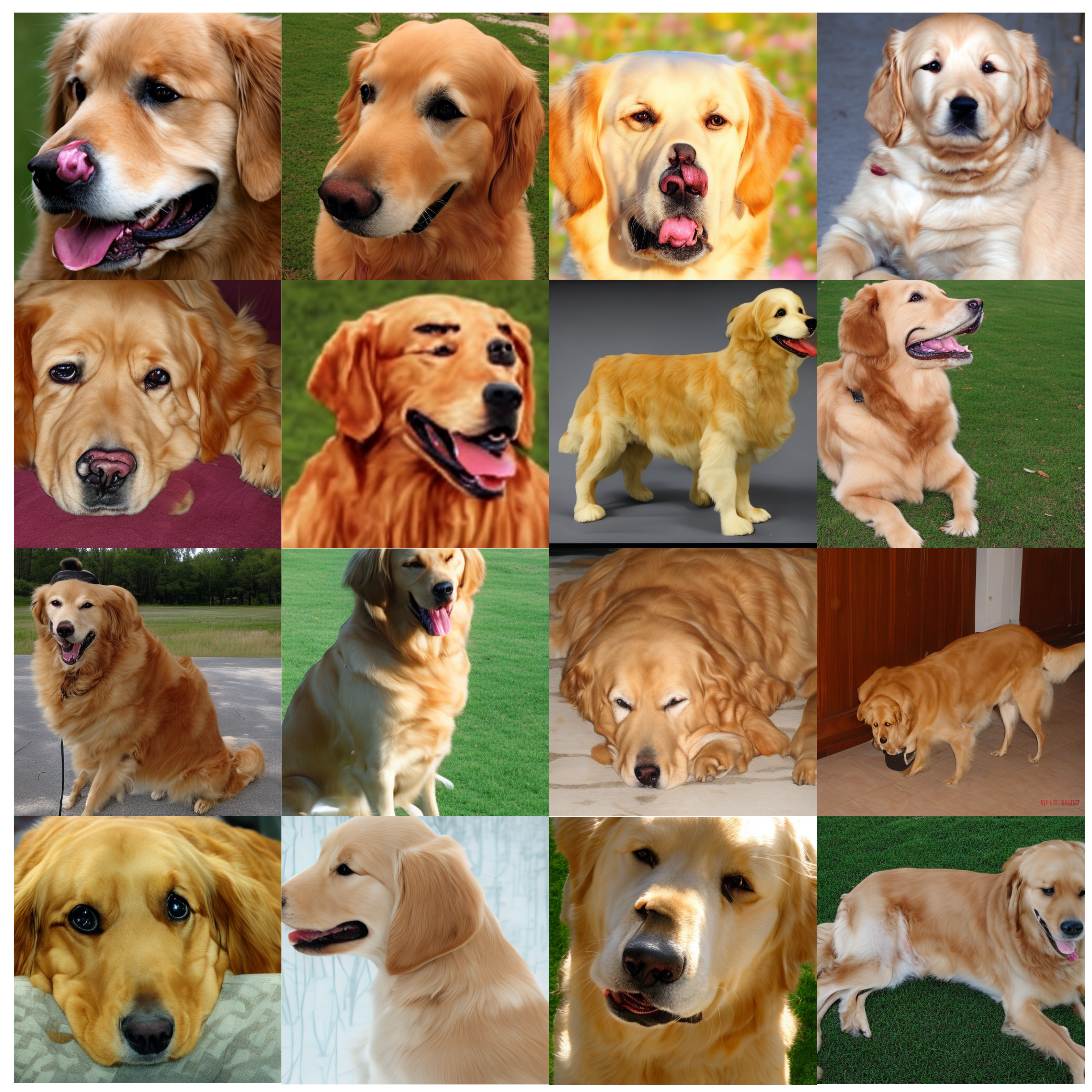}
    \caption{Uncurated \textbf{Zero-shot Resolution Extrapolation}. $w_{\mathrm{cfg}}=3.0$, $w_{\mathrm{scg}}=2.5$. Golden retriever (ImageNet class 207).}
\label{fig:uncurated_resolution_207}
\end{figure}

\begin{figure}
\centering\includegraphics[width=0.99\linewidth]{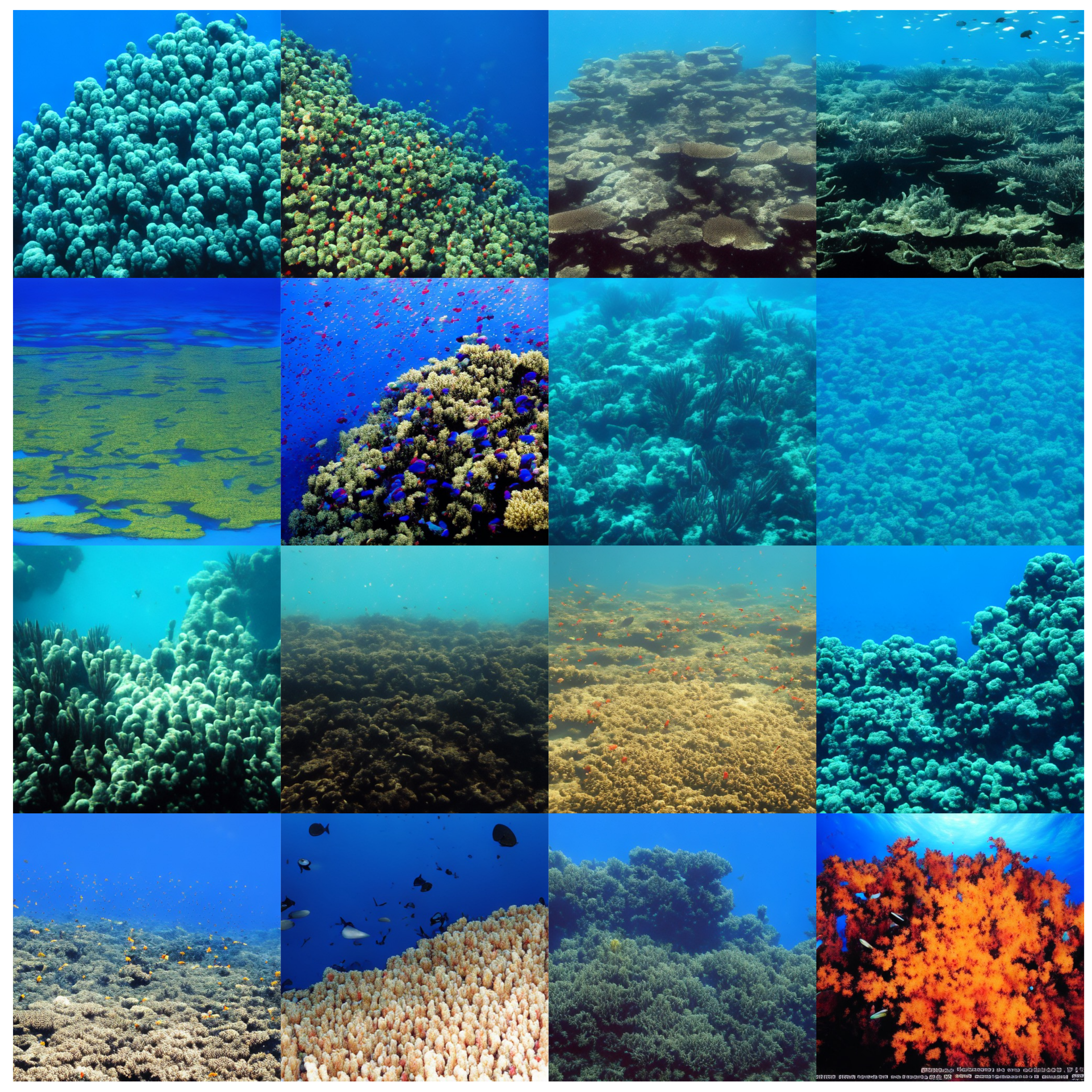}
    \caption{Uncurated \textbf{Zero-shot Resolution Extrapolation}. $w_{\mathrm{cfg}}=3.0$, $w_{\mathrm{scg}}=2.5$. Coral reef (ImageNet class 973).}
\label{fig:uncurated_resolution_973}
\end{figure}

\begin{figure}
\centering\includegraphics[width=0.99\linewidth]{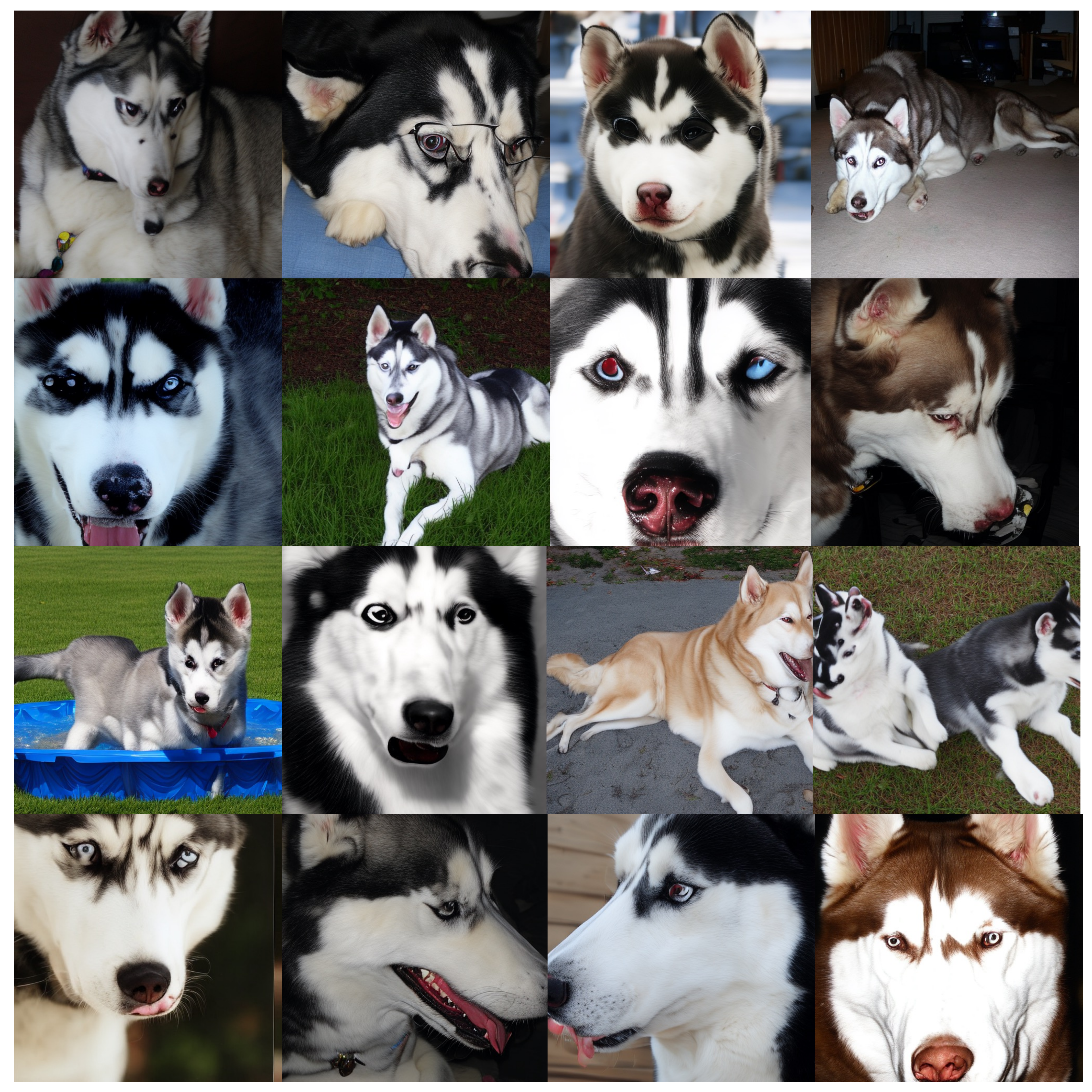}
    \caption{Uncurated \textbf{Zero-shot Resolution Extrapolation}. $w_{\mathrm{cfg}}=3.0$, $w_{\mathrm{scg}}=2.5$. Husky (ImageNet class 250).}
\label{fig:uncurated_resolution_250}
\end{figure}

\begin{figure}
\centering\includegraphics[width=0.99\linewidth]{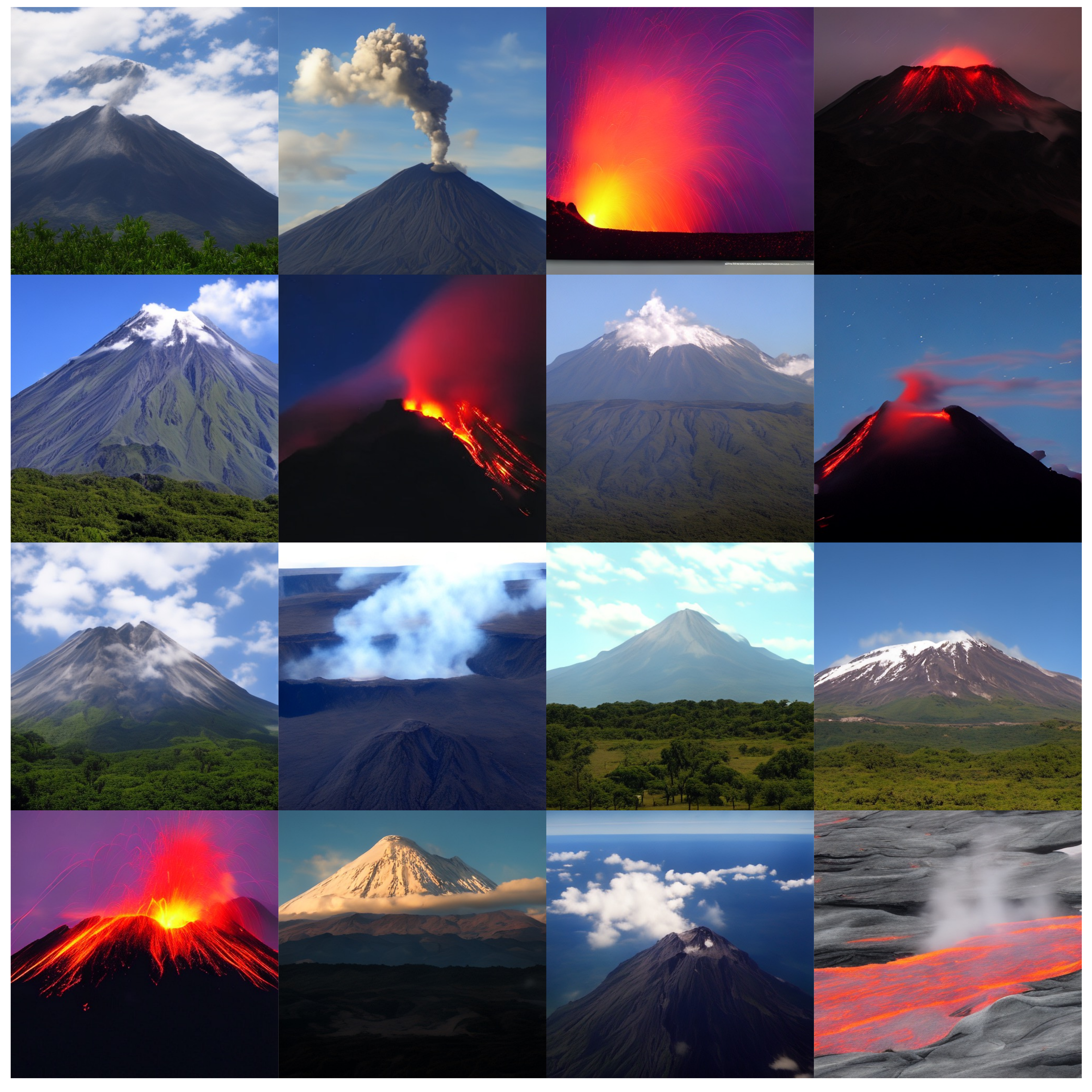}
    \caption{Uncurated \textbf{Zero-shot Resolution Extrapolation}. $w_{\mathrm{cfg}}=3.0$, $w_{\mathrm{scg}}=2.5$. Volcano (ImageNet class 980).}
\label{fig:uncurated_resolution_980}
\end{figure}

\begin{figure}
\centering\includegraphics[width=0.99\linewidth]{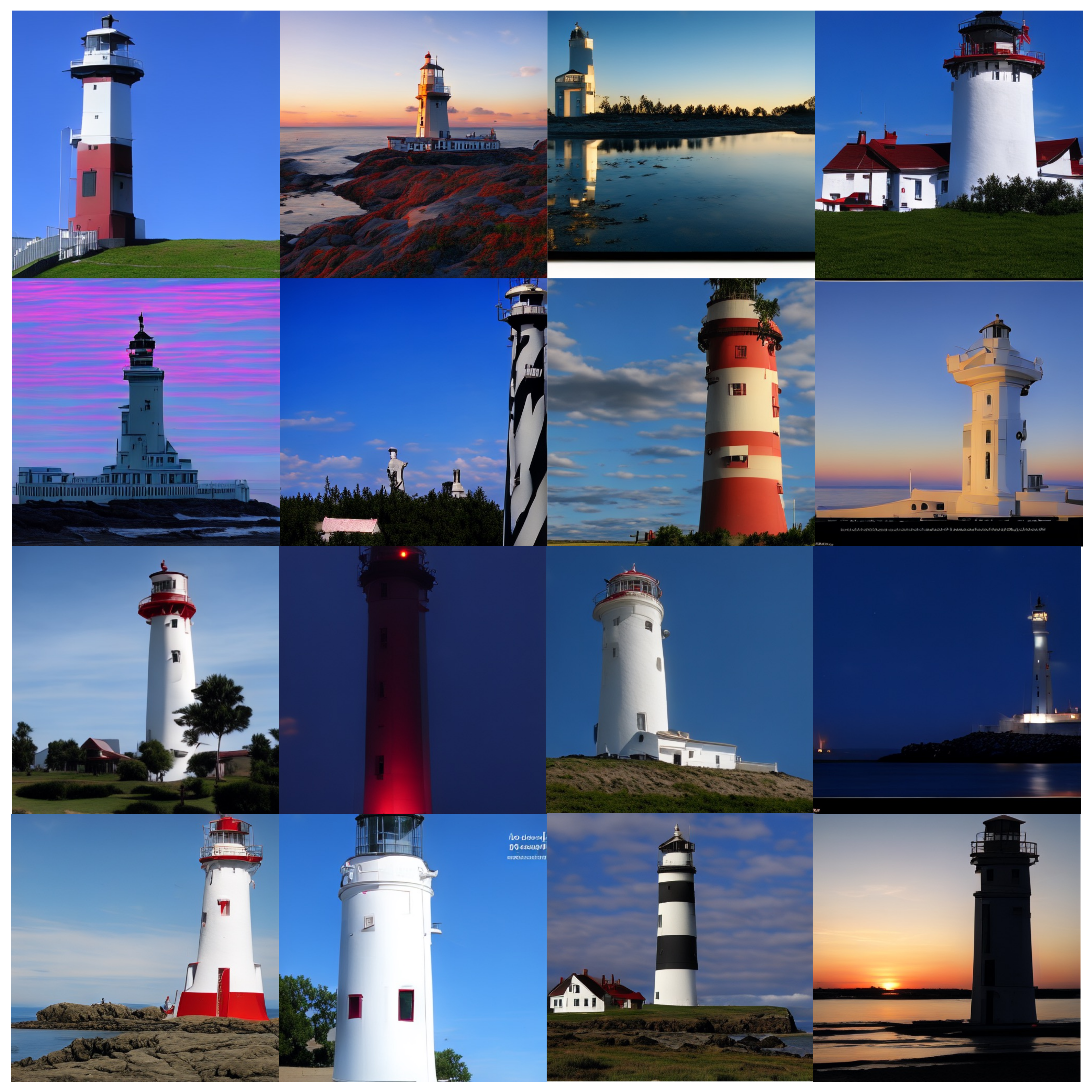}
    \caption{Uncurated \textbf{Zero-shot Resolution Extrapolation}. $w_{\mathrm{cfg}}=3.0$, $w_{\mathrm{scg}}=2.5$. Lighthouse (ImageNet class 437).}
\label{fig:uncurated_resolution_437}
\end{figure}

\begin{figure}
\centering\includegraphics[width=0.99\linewidth]{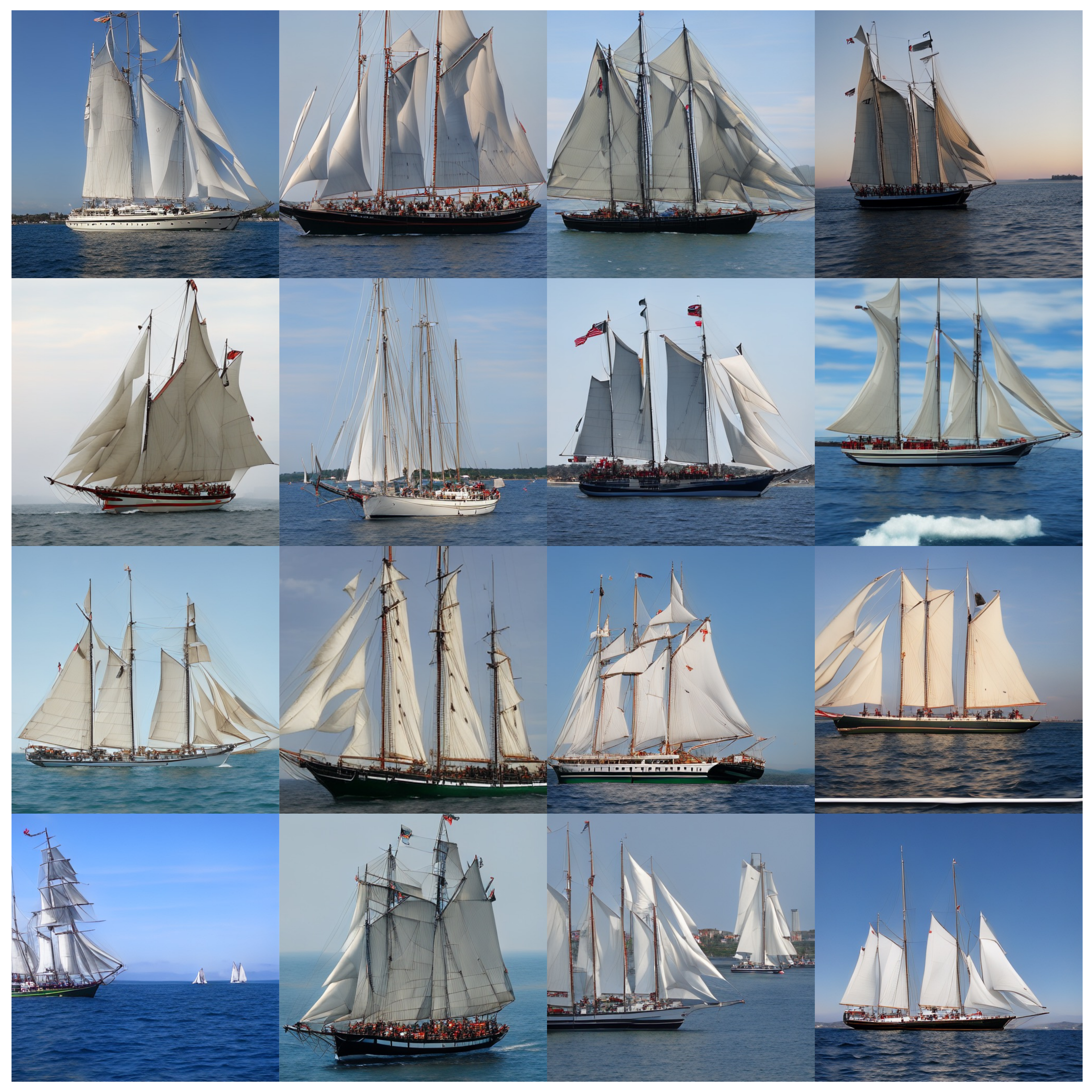}
    \caption{Uncurated \textbf{Zero-shot Resolution Extrapolation}. $w_{\mathrm{cfg}}=3.0$, $w_{\mathrm{scg}}=2.5$. Schooner (ImageNet class 780).}
\label{fig:uncurated_resolution_780}
\end{figure}

We provide uncurated resolution extrapolation results from Fig.~\ref{fig:uncurated_resolution_207} to Fig.~\ref{fig:uncurated_resolution_780} with 775M RandAR-XL. Our zero-shot extrapolation produces high-quality images with unified layouts and detailed patterns like furs of dogs (Fig.~\ref{fig:uncurated_resolution_207}), coral reefs (Fig.~\ref{fig:uncurated_resolution_973}), and scenery (Fig.~\ref{fig:uncurated_resolution_980}). However, we also notice that zero-shot resolution extrapolation is a challenging task. As the model has never been trained on high-frequency details, it will struggle with the small patterns, \emph{e.g.}, eyes and noses of dogs (Fig.~\ref{fig:uncurated_resolution_207}, Fig.~\ref{fig:uncurated_resolution_250}) and straight shapes of man-made objects (Fig.~\ref{fig:uncurated_resolution_437}).

\clearpage

\clearpage
{
    \small
    \bibliographystyle{ieeenat_fullname}
    \bibliography{main}
}


\end{document}